\documentclass[11pt]{article}

\usepackage[final]{acl}

\usepackage{microtype}

\usepackage[english,bidi=default]{babel} 

\babelfont{rm}[
  BoldFont=TeXGyreTermesX-Bold.otf,
  ItalicFont=TeXGyreTermesX-Italic.otf,
  BoldItalicFont=TeXGyreTermesX-BoldItalic.otf
]{TeXGyreTermesX-Regular.otf}

\usepackage{graphicx}
\usepackage{booktabs,tabularx,siunitx}
\usepackage{amsmath} 
\usepackage[most]{tcolorbox}
\usepackage{makecell}
\usepackage{multirow} 
\usepackage{array}

\usepackage{amssymb}  
\newcommand{\cmark}{\textcolor{green!60!black}{\checkmark}}
\newcommand{\xmark}{\textcolor{red!70!black}{$\times$}}

\makeatletter
\newcommand{\GTReleaseRepo}{%
\ifacl@anonymize%
\url{https://anonymous.4open.science/r/cs-annotation-data-0231}%
\else%
\url{https://github.com/gold-vs-translation/cs-annotation-data}%
\fi%
}

\newcommand{\GTAnalysisRepo}{%
\ifacl@anonymize%
\url{https://anonymous.4open.science/r/annotation-and-analysis-scripts-0A89}%
\else%
\url{https://github.com/gold-vs-translation/annotation-and-analysis-scripts}%
\fi%
}

\newcommand{\GTDataProcessingRepo}{%
\ifacl@anonymize%
\url{https://anonymous.4open.science/r/data-processing-3D19}%
\else%
\url{https://github.com/gold-vs-translation/data-processing}%
\fi%
}

\newcommand{\GTHarnessRepo}{%
\ifacl@anonymize%
\url{https://anonymous.4open.science/r/lm-evaluation-harness-5E5A}%
\else%
\url{https://github.com/gold-vs-translation/lm-evaluation-harness}%
\fi%
}
\makeatother

\title{Gold vs.~Translation: Are We Measuring Language, Knowledge, or Locality in Translated Benchmarks?}

\author{
  \textbf{Klaudia-Doris Thellmann\textsuperscript{1}},
  \textbf{Bernhard Stadler\textsuperscript{1}}\\
  \textbf{Dan Saattrup Nielsen\textsuperscript{2}},
  \textbf{Michael Färber\textsuperscript{1}},
  \textbf{Jens Lehmann\textsuperscript{1}}\\
  \textsuperscript{1}{firstname.lastname@tu-dresden.de}
  \textsuperscript{2}{dan.smart@alexandra.dk}
\\
\\
  \textsuperscript{1}TUD Dresden University of Technology and ScaDS.AI,\\
  \textsuperscript{2}Alexandra Institute
\\
}

\begin{document}

\maketitle
\begin{abstract}
Translated versions of English-origin benchmarks such as MMLU are widely used to evaluate large language models in non-English languages, yet it remains unclear whether they measure the same constructs as natively created benchmarks.
We frame this as a \emph{proxy validity} question and compare DeepL-translated MMLU proxies against six native benchmarks in seven pairings across five Nordic and Baltic languages, evaluating 29 models on each pairing.
Beyond ranking agreement, we test \emph{incremental} proxy validity: does the translated score add predictive value for native performance after controlling for model size and language proficiency?
Proxy validity varies substantially across benchmark pairings.
Applying the same robustness criteria to all seven pairings, we find four robust effects, two marginal effects, and one null result.
The strongest evidence occurs for curriculum-overlap benchmarks such as Skolprov ($\Delta R^2=.128$, 95\% CI $[.038,.253]$), whereas translated MMLU provides no significant incremental value for Sweden-specific factual knowledge in SwedishFacts ($p=.242$).
Because SwedishFacts is also the only open-ended benchmark, we cannot fully separate the effects of local content and response format.
Overall, the results indicate a clear pattern: translated benchmarks provide a better proxy when their source content overlaps with the native evaluation target, while locally anchored knowledge is less reliably captured.
\end{abstract}

\section{Introduction}
\label{sec:intro}

Multilingual evaluation of large language models (LLMs) frequently relies on translated versions of English-origin benchmarks.
Datasets such as MMLU \citep{hendrycks2021measuring} are routinely machine-translated into dozens of languages and used to assess model performance as if they were native evaluation instruments in the target language.
Yet a growing body of evidence suggests that this practice is problematic.
Translated benchmarks carry cultural and linguistic biases from the source language \citep{Singh_2025}, exhibit translation artifacts that can inflate scores and alter system rankings \citep{Zhang_2019,Wang_2023}, and correlate less with human judgments than natively developed benchmarks \citep{Wu_2025}.

These findings raise a fundamental question: when we evaluate a model on MMLU translated into Estonian or Swedish, what do we actually measure -- the model's language proficiency, its general knowledge, or its familiarity with locally relevant content?

Most prior work addresses the \emph{channel question}: is machine translation a good enough substitute for human translation \citep{Rajaee_2025,Vroe_2025,Agrawal_2024}?
A few efforts go further: \citet{Wu_2025} find that localized benchmarks correlate more with human preferences than translated ones, and \citet{Hupkes_2025_MultiLoKo} quantify a substantial Locality Effect.
However, these studies remain descriptive. 
They document that gaps exist but do not control for the confound that strong models score well on any benchmark, nor formalize when a translated benchmark is a \emph{valid proxy} for a native one.

We address this gap by framing the problem as a \textbf{proxy validity} question: does a translated benchmark lead to the same evaluation conclusions as a native one, once we account for general model strength?
We compare translated MMLU proxies against six native benchmark sources, yielding seven native--proxy pairings across five Nordic and Baltic languages and evaluating 29 LLMs.
Beyond ranking agreement, we test \emph{incremental} proxy validity---whether the proxy score adds predictive value for native performance after controlling for model size and language proficiency---and examine content differences through culturally sensitive (CS) vs.\ agnostic (CA) stratification and required-knowledge annotations.

Our analysis reveals that proxy validity varies: ranking agreement ranges from $\tau_b=0.89$ for curriculum-overlap benchmarks to $0.51$ for Sweden-specific factual knowledge.
Under the same evidence definition for all seven pairings, four show robust incremental proxy validity, two are marginal, and one is null.
The strongest evidence occurs for Skolprov ($\Delta R^2=.128$, 95\% CI $[.038,.253]$), whereas translated MMLU provides no significant incremental value for SwedishFacts ($p=.242$), potentially due to its Sweden-centric content and/or its open-ended generation format.
Adding English MMLU as a control supports the same overall pattern, indicating that general MMLU-content familiarity alone does not explain the proxy-validity differences.
Taken together, these results indicate that the usefulness of a translated benchmark depends on how closely its content matches the knowledge targeted by the native benchmark: broadly shared curriculum content transfers well, whereas locally specific knowledge is less reliably captured.

Our contributions are:

(1)~an \emph{incremental proxy validity framework} that goes beyond rank correlation by controlling for confounds and testing whether translated benchmarks carry predictive signal beyond background factors;

(2)~a multilingual cultural-sensitivity and required-knowledge annotation pipeline validated against three-annotator majority references for 500 items across five languages;

(3)~empirical evidence across seven native-vs-proxy pairings showing that proxy validity varies with content and locality overlap, offering practical guidance on when translated benchmarks approximate native evaluation signal and when native benchmarks remain necessary.

We release the annotations, annotation pipeline, and analysis code to support reproducibility and extension to other languages and translation systems.%
\footnote{Annotations and merged analysis data: \GTReleaseRepo; analysis scripts and annotation pipeline: \GTAnalysisRepo.}

\section{Related Work}
\label{sec:rel_work}

\paragraph{Translation methods.}
Studies comparing MT and HT benchmarks find high rank correlations on average, but also reveal systematic overestimation for commonly benchmarked low-resource languages~\citep{Rajaee_2025}, disproportionate effects on low-resource languages~\citep{Agrawal_2024}, sensitivity to model type~\citep{Vroe_2025}, and substantial cross-lingual gaps~\citep{Ahuja_2023,Lai_2023}. 
Post-editing~\citep{Huang_2025} and adapted translate-test~\citep{Artetxe_2023} mitigate but do not eliminate these effects.
All of these compare translation \emph{methods} on English-origin content; we ask whether \emph{any} translated version measures the same construct as a native L2 benchmark.

\paragraph{Native versus translated benchmarks.}
\citet{Wu_2025} find that localized benchmarks correlate more with human preferences than translated ones; \citet{Singh_2025} show that rankings shift on culturally sensitive items in GlobalMMLU; \citet{Keleg_2023} show culturally patterned factual knowledge in pretrained LMs; and \citet{Hupkes_2025_MultiLoKo} quantify a substantial Locality Effect across 31 languages.
However, none of these formalize proxy validity with statistical controls.
We build on these findings by testing whether the divergence persists after controlling for confounds.

\paragraph{Translationese.}
Translation artifacts inflate scores, alter rankings \citep{Zhang_2019,Wang_2023}, and create spurious cross-lingual performance gains \citep{Park_2024,Plazaortiz_2025}; we ask whether the net effect undermines proxy validity.

\paragraph{Validity frameworks.}
Recent work has called for more rigorous evaluation methodology: \citet{Salaudeen_2025} propose a validity-centered framework distinguishing different types of validity threats in AI evaluation, \citet{Bean_2025} find widespread construct validity threats across 445 LLM benchmarks, and \citet{Alaa_2025} show that high benchmark scores do not predict real-world performance.
More broadly, \citet{Wallach_2025} and \citet{Weidinger_2025} argue that evaluating GenAI requires social science measurement rigor.
On the empirical side, LLM capabilities are non-monolithic with a strong general ability factor \citep{Burnell_2023,Ilic_2024_AGA,Kim_2025_BenchmarkProfiling} -- motivating our use of controls to separate general strength from benchmark-specific signal.
Our notion of proxy validity is distinct from construct validity as discussed in these works: we ask whether one \emph{observable} benchmark (translated) can substitute for another \emph{observable} benchmark (native), rather than whether a benchmark measures a latent construct (see Appendix~\ref{sec:app:related} for extended discussion).

\section{Datasets and Comparison Setup}
\label{sec:datasets}

In this section we introduce the native benchmark collection used in our study and present the pairings that match each native benchmark against a translated MMLU proxy in the same language.

\begin{table}[htbp]
  \centering
  \footnotesize
  \renewcommand{\arraystretch}{1.05}
  \begin{tabularx}{\columnwidth}{l c c c}
    \toprule
    \textbf{Native (Gold)} & \textbf{$n$} & \textbf{MMLU-Proxy} & \textbf{$n$} \\
    \midrule
    Exam-ET            & 1{,}614 & ET & 2{,}800 \\
    INCLUDE-ET          & 224     & ET & 2{,}800 \\
    Skolprov            & 545     & SV & 2{,}800 \\
    SwedishFacts        & 1{,}292 & SV & 2{,}800 \\
    DanishCitizenTest   & 668     & DA & 2{,}800 \\
    NRK-Quiz         & 3{,}600 & NB & 2{,}701 \\
    INCLUDE-LT          & 534     & LT & 2{,}800 \\
    \bottomrule
  \end{tabularx}
  \caption{
  Dataset overview and native-vs-proxy comparison pairings. 
  Each native benchmark is paired with the DeepL-translated MMLU in the same language (G-MMLUX subset). 
  Proxies sourced from EU20-MMLU for ET, LT, SV, DA and from m\_mmlu for NB.
  }
  \label{tab:overview_comparison-pairings}
\end{table}

Our study compares \emph{translated} MMLU proxies against \emph{natively created} benchmarks across five target languages: Estonian~(ET), Swedish~(SV), Danish~(DA), Norwegian Bokmål~(NB), and Lithuanian~(LT).
Table~\ref{tab:overview_comparison-pairings} summarizes all datasets used in our study, organized into \emph{translated global benchmarks} derived from MMLU and \emph{native local benchmarks} created independently in the target language.

\subsection{Translated Benchmarks (Proxies)}
Our MMLU proxies are drawn from EU20-MMLU \citep{thellmann2024multilingual} for ET, LT, SV, and DA, and from m\_mmlu \citep{alexandrainst_m_mmlu} for NB---both DeepL translations of the full MMLU \citep{hendrycks2021measuring}.
We additionally construct \textbf{G-MMLUX}, a unified subset mapping the culturally annotated items from GlobalMMLU \citep{Singh_2025} to their translations in each target language, including the English originals.
After agreement filtering to exclude items with low annotator consensus, G-MMLUX contains 16{,}701 items across six languages (Appendix~\ref{sec:app:datasets}).

\subsection{Native Benchmarks}
\label{sec:benchmarks}

We include six natively created benchmark sources across five languages, yielding seven native sets through the independent Estonian and Lithuanian INCLUDE subsets:

\begin{description}
    \item[\textbf{Exam-ET}] \citep{lillepalu2026estonian} contains 1{,}614 MC questions from official Estonian secondary- and high-school exams across seven subjects, one of which we subdivide.
    \item[\textbf{Skolprov}] \citep{ekgren_2025_skolprov} comprises 545 MC items from the Swedish university entrance exam and professional licensing exams for physicians, dentists, audiologists, and pharmacists.
    \item[\textbf{SwedishFacts}] \citep{kunz2026diagnostic} is a generative factual QA benchmark of 1{,}292 items targeting Sweden-specific personalities and events, and is the only non-MC dataset in our study.
    \item[\textbf{DanishCitizenTest}] \citep{nielsen2024danishct,nielsen2023scandeval} consists of 668 MC items from the official Danish citizenship and residence permit exams.
    \item[\textbf{NRK-Quiz}] \citep{mikhailov2025collection} provides 3{,}600 MC quiz questions from NRK, Norway's public broadcaster, covering Norwegian culture, geography, and current affairs.
    \item[\textbf{INCLUDE}] \citep{romanou2024include} contributes the Estonian (224 items) and Lithuanian (534 items) subsets of natively sourced exam questions.
\end{description}

All native sets were created in the target language and are free of translation artifacts by construction, but they differ markedly in locality and content specificity.
The proportion of culturally sensitive (CS) items ranges from 14.3\% (Skolprov) to 97.8\% (SwedishFacts), compared to 27.8\% in GlobalMMLU (Table~\ref{tab:app:cultural-annotations}).

\subsection{Subject Mapping}
\label{sec:subjmapping}

To assess whether the native and translated benchmarks are thematically comparable, we mapped each native subject to the most appropriate MMLU category (\textsc{stem}, \textsc{humanities}, \textsc{social sciences}, \textsc{business}, \textsc{medical}, \textsc{other}).

A direct mapping between native subjects and individual MMLU subjects is unreliable because the underlying taxonomies differ in granularity and scope and often lack one-to-one equivalents.
We therefore conduct cross-benchmark analyses at the broader category level, which provides a more robust basis for comparison.
Across native-exam and GlobalMMLU-Lite cross-validation items, GPT-5.2 category labels agree with reference categories on 85.1\% of 3{,}044 items.
Disagreements mainly reflect MMLU taxonomy ambiguities rather than clear mapping errors.
The full mapping and examples of ambiguous cases are provided in Appendix~\ref{sec:app:datasets:mapping}.

The category mapping also reveals how much thematic ground the native benchmarks share with MMLU.
Table~\ref{tab:overview_comparison-pairings} lists our seven comparison pairings.
Exam-ET and Skolprov are curriculum-based exam benchmarks whose items can be mapped to MMLU categories at \emph{approximate} confidence for 70\% (1{,}130/1{,}614) and 85\% (465/545) of items, respectively---primarily in \textsc{stem}, \textsc{humanities}, and \textsc{medical}.
DanishCitizenTest maps entirely to \textsc{social sciences} at approximate confidence, but its 95.4\% culturally sensitive content has no meaningful MMLU counterpart.
In our analysis, INCLUDE~(ET, LT), NRK-Quiz, and SwedishFacts are classified entirely via item-level annotation at the category level, instead of using native subject labels (Appendix~\ref{sec:app:datasets:comparison}).

\section{Annotation and Meta-Evaluation}
\label{sec:annotation}

This section describes how we obtain culturally sensitive~(CS) versus culturally agnostic~(CA) labels for benchmark items and how we validate the automatic annotation pipeline.
The distinction is central to our proxy validity analysis (\S\ref{sec:proxy}), as it allows us to test whether translated benchmarks fail specifically on items requiring local or culture-specific knowledge.
We pursue a two-pronged approach:
\textbf{(1)}~a native-language human reference with three independent annotators per item for 500 items across five languages, and
\textbf{(2)}~a GPT-5.2 annotation pipeline calibrated on GlobalMMLU-Lite.

\subsection{Human Annotation}
\label{sec:annotation:human}

We adapted the GlobalMMLU cultural-sensitivity schema \citep{Singh_2025} for our native benchmark items.
For each item, annotators record a primary \texttt{required\_knowledge} type (\textit{none}, \textit{cultural}, \textit{regional}, or \textit{dialect}), a binary \texttt{time\_sensitive} flag, an explicit \texttt{cultural\_sensitivity\_label} (CS/CA), a \texttt{subject\_category}, and optional metadata including trigger spans and geographic/cultural identifiers.

The CS/CA label captures whether annotators judge an item to require culture-specific, regional, or dialect knowledge, but it is recorded as a separate binary judgment rather than mechanically derived from \texttt{required\_knowledge}.
For 10 of the 500 items (2\%), the explicit CS/CA majority label differs from the label one would obtain by mechanically collapsing \texttt{required\_knowledge} into \texttt{none} versus non-\texttt{none}.
For each target language, we sampled 100 native-benchmark items and obtained three independent professional annotations per item, yielding 1{,}500 raw ratings over 500 items.
Annotators were recruited based on native-speaker status and professional language expertise.
For the reliability and meta-evaluation analyses, we use the original raw/pre-QA human annotations as the reference labels, before any GPT-assisted consistency review.
The full schema is given in Appendix~\ref{sec:app:annotation:guide}.
Sampling, recruitment, and raw/pre-QA reconstruction details are documented in Appendix~\ref{sec:app:annotation:details}.

Table~\ref{tab:annot-label-dist} reports the resulting three-annotator majority reference.
Explicit CS/CA has a unique majority for all 500 items.
For four-way \texttt{required\_knowledge}, 477 items have a unique majority.
The remaining 23 items (4.6\%) are unresolved 1--1--1 ties, most frequently in Lithuanian (13 of 100).

\begin{table}[htbp]
\centering
\footnotesize
\setlength{\tabcolsep}{0.32em}
\renewcommand{\arraystretch}{1.05}
\begin{tabular}{@{}lrrrrrrr@{}}
\toprule
& \multicolumn{5}{c}{\textbf{Required knowledge}} 
& \multicolumn{2}{c}{\textbf{Explicit CS/CA}} \\
\cmidrule(lr){2-6}\cmidrule(l){7-8}
\textbf{L} 
& \texttt{none} 
& \texttt{reg.} 
& \texttt{cult.} 
& \texttt{dial.} 
& \textbf{ties}
& \textbf{CS}
& \textbf{CA} \\
\midrule
DA &  4 & 85 &  6 & 0 &  5 & 97 &  3 \\
ET & 77 & 19 &  3 & 1 &  0 & 22 & 78 \\
LT & 34 & 41 & 12 & 0 & 13 & 66 & 34 \\
NB & 23 & 30 & 41 & 2 &  4 & 75 & 25 \\
SV & 45 &  2 & 52 & 0 &  1 & 59 & 41 \\
\midrule
\textbf{All} 
& \textbf{183} 
& \textbf{177} 
& \textbf{114} 
& \textbf{3} 
& \textbf{23}
& \textbf{319}
& \textbf{181} \\
\bottomrule
\end{tabular}
\caption{
Three-annotator majority reference for 100 native items per language (500 total).
Required knowledge and explicit CS/CA (culturally sensitive / culturally agnostic) are independently annotated.
Required-knowledge counts exclude 23 unresolved 1--1--1 ties.
Explicit CS/CA has a unique majority for all 500 items.
}
\label{tab:annot-label-dist}
\end{table}

Inter-annotator agreement on the explicit CS/CA field is generally high but varies across languages.
Unanimous 3-of-3 agreement ranges from 65\% in Norwegian Bokm{\aa}l to 91\% in Estonian. 
Fleiss'~$\kappa$ ranges from .25 in Danish to .83 in Estonian, and Gwet's AC1 from .60 in Norwegian Bokm{\aa}l to .91 in Estonian.
The Danish case illustrates the prevalence effect: although Fleiss'~$\kappa$ is low ($.25$, 95\% CI $[-.05,.51]$), annotators still reach 84\% unanimous agreement and AC1=.88 because almost all raw ratings are CS.
Full multilingual IAA statistics, required-knowledge agreement, and bootstrap CIs are reported in Appendix~\ref{sec:app:annotation:iaa}; annotation-provenance details in~\ref{sec:app:annotation:details}.

\subsection{Automatic Annotation Pipeline}
\label{sec:annotation:automatic}
To scale annotation to the full native benchmark collection (Table~\ref{tab:overview_comparison-pairings}), we use GPT-5.2 with structured JSON output.
The pipeline assigns the same GlobalMMLU-style labels as the human annotators;
Appendix~\ref{sec:app:annotation:prompt} reproduces the full prompt in Figure~\ref{fig:annotation-prompt}.
We calibrated the prompt on GlobalMMLU-Lite, which provides 3--8 independent annotations per item (median:~4).
We aggregate \texttt{required\_knowledge} by majority vote, resolving simple ties between \textit{none} and one non-none label in favor of the non-none category and discarding complex ties with multiple non-none labels.
This yields 391 usable items (of 400; 226~\textit{none}, 48~\textit{cultural}, 114~\textit{regional}, 3~\textit{dialect}).

For GlobalMMLU-Lite only, the CS/CA reference follows its documented rule: CA if strictly more than half of annotators assigned \textit{none}, and CS otherwise, resulting in 191~CS and 200~CA items.
From the 391 usable items, we created a stratified dev/test split (174/217 items).
Iterative refinement on the dev split addressed five recurring error patterns, including under-detection of Western philosophical terminology and implicit US-centric regional cues.
Details are provided in Appendix~\ref{sec:app:annotation:error-patterns}.

\subsection{Meta-Evaluation}
\label{sec:annotation:metaeval}

We evaluate the annotation pipeline against two reference sets:
\textbf{(1)}~the held-out GlobalMMLU-Lite test split ($n=217$), and
\textbf{(2)}~the Native-500 reference described above.
Both evaluations cover binary CS/CA classification and four-way \texttt{required\_knowledge} classification.
For Native-500, the binary CS/CA evaluation uses the explicit three-annotator majority label for all 500 items, whereas the four-way \texttt{required\_knowledge} evaluation uses the 477 items with a unique majority.
The prompt was calibrated only on GlobalMMLU-Lite. 
Native-500 was not used for prompt tuning, and GPT annotations for the native items were joined to the human-majority reference only for this meta-evaluation (Appendix~\ref{sec:app:annotation:metaeval}).

\begin{table}[htbp]
  \centering
  \footnotesize
  \setlength{\tabcolsep}{0.35em}
  \renewcommand{\arraystretch}{1.05}
  \begin{tabular}{l c c c c c c}
    \toprule
    & \textbf{n} & \textbf{Acc.} & \multicolumn{2}{c}{\textbf{CS}} & \multicolumn{2}{c}{\textbf{CA}} \\
    \cmidrule(lr){4-5} \cmidrule(lr){6-7}
    & & & P & R & P & R \\
    \midrule
    GlobalMMLU-Lite & 217 & 73.3 & 79.5 & 61.7 & 69.4 & 84.5 \\
    Native (5~langs) & 500 & 90.6 & 89.8 & 96.2 & 92.4 & 80.7 \\
    \bottomrule
  \end{tabular}
  \caption{
  Binary CS/CA agreement between GPT-5.2 and reference labels (\%).
  CS/CA denotes culturally sensitive / culturally agnostic.
  For GlobalMMLU-Lite, the CS/CA reference follows the documented rule over human \texttt{required\_knowledge} votes.
  For Native-500, the reference is the majority of three independent explicit CS/CA annotations per item.
  }
  \label{tab:meta-eval-csca}
\end{table}

\paragraph{CS/CA agreement.}
The pipeline identifies binary CS/CA labels more reliably on Native-500 than on the held-out GlobalMMLU-Lite test split.
Table~\ref{tab:meta-eval-csca} reports binary CS/CA performance.
On GlobalMMLU-Lite, GPT-5.2 reaches 73.3\% accuracy, with lower recall for CS items (61.7\%) than for CA items (84.5\%), indicating that implicit cultural cues in English-origin items remain difficult.
Against the explicit three-annotator native majority, accuracy rises to \textbf{90.6\%} (macro-F1 89.5\%).
The native error profile is asymmetric: 
CS recall is high (96.2\%), whereas CA recall is lower (80.7\%), so remaining errors more often over-predict cultural sensitivity than miss it.
As sensitivity checks, deriving GPT CS/CA from predicted \texttt{required\_knowledge} yields 92.8\% accuracy against the explicit native majority; deriving CS/CA from \texttt{required\_knowledge} on both sides yields 94.4\% (Appendix~\ref{sec:app:annotation:metaeval}).

\begin{table}[htbp]
  \centering
  \footnotesize
  \setlength{\tabcolsep}{0.82em}
  \begin{tabular}{@{}llrrr@{}}
    \toprule
    \textbf{Set} & \textbf{Class} & \textbf{P} & \textbf{R} & \textbf{F1} \\
    \midrule
    \multirow{4}{*}{\shortstack[l]{GlobalMMLU\\($n=217$)}}
      & \textit{cultural} & 59.5 & 75.8 & 66.7 \\
      & \textit{regional} & 82.5 & 55.9 & 66.7 \\
      & \textit{none}     & 76.1 & 82.9 & 79.4 \\
      & \textit{dialect}  & 100\phantom{.0} & 50.0 & 66.7 \\
    \midrule
    \multirow{4}{*}{\shortstack[l]{Native\\($n=477$)}}
      & \textit{cultural} & 78.6 & 57.9 & 66.7 \\
      & \textit{regional} & 76.4 & 93.2 & 84.0 \\
      & \textit{none}     & 96.0 & 90.7 & 93.3 \\
      & \textit{dialect}  & 75.0 & 100\phantom{.0} & 85.7 \\
    \bottomrule
  \end{tabular}
  \caption{
    Per-class precision (P), recall (R), and F1 (\%) for \texttt{required\_knowledge} predictions by GPT-5.2.
    Native metrics are computed on the 477 items with a unique three-annotator majority.
    Confusion matrices and tie handling are in Appendix~\ref{sec:app:annotation:metaeval}.
  }
  \label{tab:meta-eval-reqknow}
\end{table}

\paragraph{Required knowledge.}
Table~\ref{tab:meta-eval-reqknow} reports per-class precision, recall, and F1 for the four-way task.
Although harder than binary CS/CA classification, performance remains strong for the main classes.
On GlobalMMLU-Lite, \textit{regional} recall is comparatively low (55.9\%), reflecting confusion between implicit regional cues and \textit{none}.
On Native-500, the primary evaluation uses the 477 items with a unique human majority.
Accuracy is \textbf{83.9\%} and macro-F1 is 82.4\%.
If the 23 unresolved human ties are conservatively counted as errors, accuracy is 80.0\% (400/500).
The dominant native error is confusion from \textit{cultural} to \textit{regional}, while \textit{none} and \textit{regional} achieve F1 scores above 83\%.
Dialect scores are based on only three resolved native items and should be treated with caution.

\section{Proxy Validity Evaluation}
\label{sec:proxy}
In this section we test whether translated MMLU proxies produce the same model rankings as native benchmarks (\S\ref{sec:proxy:ranking}), 
then ask the stronger validity question: whether translated scores add predictive signal beyond model size and language proficiency (\S\ref{sec:proxy:incremental}).
We quantify uncertainty by bootstrapping models and test model-sample stability with leave-one-family-out (LOFO) refits and an English-MMLU control sensitivity (\S\ref{sec:proxy:robustness}).

\subsection{Evaluation Setup}
\label{sec:proxy:setup}
We evaluate 29 instruction-tuned LLMs spanning four size categories: M ($\geq$70B, 7~models), S (15--32B, 8~models), XS (7--9B, 7~models), and XXS ($\leq$1.7B, 7~models), covering model families including Llama, Qwen, Gemma, Mistral, Falcon, and several European-focused models (Appendix~\ref{sec:app:proxy:models}).
All evaluations use the LM Evaluation Harness \citep{eval-harness} with custom task implementations for the native benchmarks.
We compute accuracy scores on all native benchmarks and their MMLU proxies (Table~\ref{tab:overview_comparison-pairings}), as well as on Belebele \citep{bandarkar2024belebele} in all five target languages as a language-proficiency control.
Per-model outputs are merged with the CS/CA annotations from \S\ref{sec:annotation} for the stratified secondary analyses (Appendix~\ref{sec:app:proxy:data}).

\subsection{Ranking Analysis}
\label{sec:proxy:ranking}
We assess ranking consistency between native benchmarks and their MMLU proxies using Kendall's~$\tau_b$, reporting 95\% bootstrap confidence intervals and top-5 overlap (Appendix~\ref{sec:app:proxy:blockb}).
As a preliminary check, CA and CS model rankings within the translated MMLU proxies are highly stable across languages ($\tau_b=.91$--$.94$; Appendix~\ref{sec:app:proxy:blocka}).

\begin{table}[htbp]
  \centering
  \small
  \setlength{\tabcolsep}{0.7em}
  \renewcommand{\arraystretch}{1.05}
  \begin{tabular}{@{}llcccc@{}}
    \toprule
    \textbf{Native} & \textbf{L} & \textbf{$\tau_b$\,[95\%\,CI]} & \textbf{T5} & $\boldsymbol{\tau}_\textbf{CA}$ & $\boldsymbol{\tau}_\textbf{CS}$ \\
    \midrule
    Skolprov       & SV & .89\,[.78,\,.96] & 5/5 & .88 & .77 \\
    Exam-ET        & ET & .88\,[.78,\,.96] & 5/5 & .83 & .88 \\
    INCLUDE        & LT & .87\,[.79,\,.93] & 5/5 & .82 & .85 \\
    NRK-Quiz       & NB & .85\,[.76,\,.92] & 4/5 & .82 & .88 \\
    DanishCT       & DA & .82\,[.73,\,.90] & 3/5 & .66 & .84 \\
    INCLUDE        & ET & .72\,[.58,\,.83] & 3/5 & .68 & .72 \\
    SwedishFacts   & SV & .51\,[.25,\,.74] & 3/5 & .50 & .52 \\
    \bottomrule
  \end{tabular}
  \caption{
  Native-vs-proxy ranking consistency. 
  $\tau_b$: Kendall's~$\tau_b$ on overall accuracy across 29~models. 
  \textbf{T5}: top-5 model overlap. 
  $\tau_\text{CA}$/$\tau_\text{CS}$: $\tau_b$ restricted to culturally agnostic / sensitive items. 
  All $p<.001$. 
  Full statistics in Appendix~\ref{sec:app:proxy:blockb}.
  }
  \label{tab:proxy-rank-b}
\end{table}

Table~\ref{tab:proxy-rank-b} presents the native-vs-proxy ranking results.
Three patterns emerge.

\textbf{First, benchmarks with broad curriculum or content overlap show the strongest ranking agreement.}
Skolprov ($\tau_b = .89$), Exam-ET ($.88$), and INCLUDE-LT ($.87$) all achieve top-5 overlaps of 5/5, meaning that the five best-performing models are identical on the native benchmark and its same-language MMLU proxy.

\textbf{Second, high-CS benchmarks can still show substantial ranking agreement.}
NRK-Quiz ($\tau_b=.85$) and DanishCitizenTest ($.82$) correlate strongly with their translated proxies despite substantial local or culturally specific content.
This already suggests that CS/CA status alone does not determine ranking agreement.

\textbf{Third, SwedishFacts shows the weakest ranking agreement.}
Its correlation with the Swedish MMLU proxy is much lower ($\tau_b=.51$), suggesting that Sweden-specific factual knowledge is less well captured by the translated benchmark.
INCLUDE-ET also shows weaker agreement ($.72$) than the other multiple-choice pairings.

It is the smallest multiple-choice native benchmark ($n=224$, versus 534--3{,}600 for the other MC benchmarks), which plausibly contributes to its wider rank uncertainty.
We treat this as an explanation to test rather than an identified mechanism.

Most importantly, ranking correlation alone cannot establish proxy validity.
Strong models tend to perform well on many benchmarks \citep{Ilic_2024_AGA}, so high $\tau_b$ values may reflect general model strength rather than genuine proxy signal.
We therefore next test whether translated MMLU adds predictive value beyond model size and target-language proficiency.

\subsection{Incremental Proxy Validity}
\label{sec:proxy:incremental}

To test whether the translated proxy carries information \emph{beyond} what background factors already explain, we fit two nested OLS models for each pairing, where $y$ is the native score, $s=\log(\text{total parameters})$, $b$ is the Belebele score, and $x$ is the proxy score, all in language~$L$:
\begin{align*}
\textbf{M1}\;\text{(baseline):}\quad & y_L \sim s + b_L \\
\textbf{M2}\;\text{(extended):}\quad & y_L \sim s + b_L + x_L
\end{align*}

\noindent

If Model~2 improves over Model~1 in the nested $F$-test, the proxy carries predictive information beyond the controls.
These regressions use total benchmark scores rather than individual CS/CA labels.
The annotations enter only in the content analysis and exploratory CS/CA stratification below.
Because $n=29$ models is small for regression, we pair the nested $F$-test with 5{,}000 model-level bootstrap resamples and LOFO stability checks (Appendix~\ref{sec:app:proxy:incremental}).
We classify the strength of evidence using the same criteria for all seven pairings.
A pairing is \emph{robust} if the full-sample incremental $F$-test is significant, the 95\% bootstrap CI for $\beta_\text{proxy}$ excludes zero, and $\beta_\text{proxy}$ remains positive with nominal $p<.05$ in all 17 LOFO refits.
A \emph{marginal} pairing is full-sample significant but fails at least one robustness criterion.
A \emph{null} pairing has a non-significant full-sample $F$-test and a bootstrap $\beta_\text{proxy}$ CI containing zero.
These labels describe evidence strength, not intrinsic benchmark quality. 
We do not use the $\Delta R^2$ CI as a null test, because in-sample $\Delta R^2$ is non-negative by construction.
Near-zero lower bounds instead indicate uncertainty about the size of the incremental effect.

\begin{table}[htbp]
  \centering
  \footnotesize
  \setlength{\tabcolsep}{0.20em}
  \renewcommand{\arraystretch}{1.04}
  \begin{tabular}{@{}lccc@{}}
    \toprule
    \textbf{Native} & $\boldsymbol{\Delta R^2}$ \textbf{[95\% CI]} & $\boldsymbol{\beta_\text{proxy}}$ \textbf{[95\% CI]} & \textbf{Class} \\
    \midrule
    Skolprov       & .128 [.038,.253] & 1.06 [.80,1.19] & robust \\
    INCLUDE-LT     & .076 [.021,.148] & 0.94 [.67,1.13] & robust \\
    NRK-Quiz       & .041 [.009,.087] & 0.46 [.29,.59] & robust \\
    Exam-ET        & .024 [.010,.044] & 0.52 [.34,.85] & robust \\
    INCLUDE-ET     & .059 [.000,.199] & 0.84 [$-.02$,1.30] & marginal \\
    DanishCT       & .009 [.000,.027] & 0.31 [.01,.57] & marginal \\
    SwedishFacts   & .035 [.000,.163] & 0.14 [$-.10$,.33] & null \\
    \bottomrule
  \end{tabular}
  \caption{
Incremental proxy validity on all items.
Brackets show 95\% model-bootstrap CIs.
Classes follow the fixed evidence definition in the text.
Full details are in Appendix~\ref{sec:app:proxy:incremental}.
The $\Delta R^2$ CI is not used as a null test because in-sample $\Delta R^2\geq0$ by construction.
  }
  \label{tab:incremental-main}
\end{table}

Table~\ref{tab:incremental-main} shows the main result: \textbf{four pairings are robust, two are marginal, and one is null}.
Skolprov has the strongest robust incremental signal ($\Delta R^2=.128$, 95\% CI $[.038,.253]$).
INCLUDE-LT and NRK-Quiz are also robust despite substantial culturally sensitive content, showing that high CS content alone does not imply weak proxy validity.
Exam-ET shows a smaller but robust increment ($\Delta R^2=.024$), partly because the baseline is already very strong ($R^2_\text{base}=.954$).
DanishCitizenTest is positive in the full sample but marginal under LOFO.
INCLUDE-ET is likewise marginal because its $\beta_\text{proxy}$ CI crosses zero and one family exclusion loses significance.
For SwedishFacts, the proxy adds no significant incremental signal ($p=.242$), and its bootstrap coefficient CI contains zero.
An exploratory CS/CA stratification is reported in Appendix~\ref{sec:app:proxy:csca}.

\begin{figure*}[htbp]
  \centering
  \includegraphics[width=\textwidth]{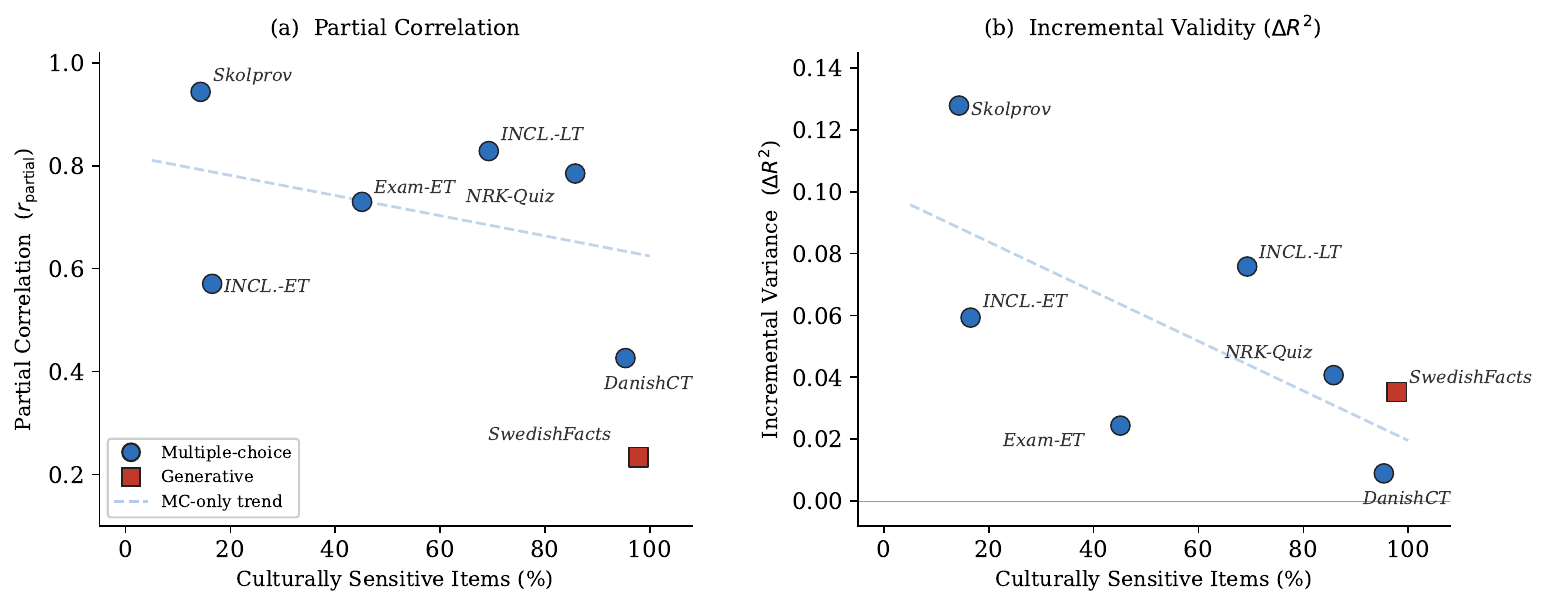}
  \caption{
  Proxy validity as a function of cultural sensitivity. 
  Each point represents one native-vs-proxy pairing. 
  \textbf{(a)}~Partial correlation after residualizing on controls. 
  \textbf{(b)}~Incremental variance ($\Delta R^2$) from the nested $F$-test. 
  The dashed line shows the MC-only linear trend (6~pairings; descriptive and not statistically significant at $n=6$). 
  DanishCitizenTest (95\% CS, MC format) already shows weak incremental validity. 
  SwedishFacts lies close to the descriptive MC extrapolation in panel~(b), which is consistent with a locality/content-overlap interpretation, but the design does not isolate locality from its open-ended response format.
  }
  \label{fig:proxy-gradient}
\end{figure*}

\subsection{Robustness}
\label{sec:proxy:robustness}

\paragraph{Leave-one-family-out sensitivity.}
To check whether the incremental effects depend on a single model family, we exclude each of 17 model families in turn and refit the main model, leaving 25--28 of the 29 models per refit.
All four robust pairings remain positive and nominally significant in every exclusion.
The two marginal pairings are less stable: 
DanishCitizenTest loses nominal significance when Cohere, Llama, or Qwen are excluded ($p=.110,.072,.096$), and INCLUDE-ET when Qwen is excluded ($p=.102$).
SwedishFacts remains non-significant in all 17 refits.
We use the LOFO $p<.05$ rule only as a conservative stability criterion for the \emph{robust} label, not as 17 multiplicity-corrected confirmatory tests (Appendix~\ref{sec:app:proxy:robustness}).

\begin{table}[htbp]
  \centering
  \setlength{\tabcolsep}{0.75em}
  \renewcommand{\arraystretch}{1.04}
  \begin{tabular}{@{}lcc@{}}
    \toprule
    \textbf{Native} & $\boldsymbol{\beta_\text{proxy}}$ \textbf{[95\% CI]} & $\boldsymbol{p_F}$ \\
    \midrule
    Skolprov       & 1.17 [.88,1.28] & $<10^{-12}$ \\
    INCLUDE-LT     & 1.08 [.84,1.30] & $<10^{-9}$ \\
    INCLUDE-ET     & 1.06 [.36,1.42] & $<10^{-5}$ \\
    NRK-Quiz       & 0.54 [.34,.68] & $<10^{-6}$ \\
    Exam-ET        & 0.59 [.39,.95] & $<10^{-7}$ \\
    DanishCT       & 0.57 [.34,.92] & $<10^{-3}$ \\
    SwedishFacts   & 0.25 [.06,.48] & .092 \\
    \bottomrule
  \end{tabular}
  \caption{
  Sensitivity analysis adding English G-MMLUX as a third control. 
  Brackets are 95\% percentile model-bootstrap CIs. 
  The main evidence classes remain those in Table~\ref{tab:incremental-main}.
  Full $\Delta R^2$, regression, and bootstrap statistics are in Appendix~\ref{sec:app:proxy:sensitivity}.
  }
  \label{tab:sensitivity-en}
\end{table}

\paragraph{Sensitivity: controlling for English MMLU performance.}
To test whether the translated-proxy association is merely an echo of general MMLU-content familiarity, we add each model's English G-MMLUX score as a third control.
Six pairings remain significant by the nested $F$-test. 
SwedishFacts remains non-significant by the $F$-test ($p=.092$), but its bootstrap CI for $\beta_\text{proxy}$ is positive ([.06,.48]), giving weak mixed evidence rather than a clean null in this alternative specification.
This does not change its \emph{null} classification under the fixed main-specification evidence definition.
The English-control analysis preserves the broad ordering but does not identify a purely language-specific causal effect.
In particular, the stronger DanishCitizenTest sensitivity result does not override its marginal classification in the main specification.

\section{Discussion}
\label{sec:discussion}

\subsection{Interpreting Proxy Validity}
\label{sec:discussion:proxy}

\paragraph{What proxy validity does and does not show.}
Our incremental test shows where translated scores predict native-benchmark performance beyond model size and target-language proficiency.
This is stronger evidence than ranking correlation alone, but it does not prove that translated and native benchmarks measure an identical latent construct or are interchangeable for every practical decision.
The robust/marginal/null labels therefore describe evidence strength for incremental association within the evaluated model set, not intrinsic benchmark quality.
Practitioners should read the results as evidence about when translated benchmarks approximate native evaluation signal, not as a blanket endorsement of translated benchmarks.

\paragraph{A content/locality-overlap gradient.}
The main pattern is a gradient: proxy validity is strongest when the native benchmark shares broad curriculum or content coverage with MMLU, and weaker or less stable as the native target becomes more locally anchored.
Skolprov and Exam-ET are robust curriculum-overlap cases.
INCLUDE-LT and NRK-Quiz also remain robust, showing that high cultural load alone does not imply poor proxy validity.
DanishCitizenTest and INCLUDE-ET are marginal, while SwedishFacts is null in the main specification.
This pattern appears across ranking analyses, incremental regressions, bootstrap uncertainty, LOFO stability, and the English-MMLU sensitivity analysis, but it is better understood as a content/locality-overlap gradient than as a deterministic function of the percentage of CS items.

\paragraph{Locality and response format.}
SwedishFacts is the clearest limitation case, but it is also the only open-ended benchmark in our study.
Its null result is consistent with a locality/content-overlap interpretation, especially because the multiple-choice pairings already vary substantially in incremental validity and DanishCitizenTest shows only a small and fragile proxy signal despite its MC format.
SwedishFacts also lies near the descriptive MC-only trend for $\Delta R^2$ in Figure~\ref{fig:proxy-gradient}.
However, the current design cannot separate Sweden-specific content from open-ended response format.
The open-ended format may amplify the mismatch, and a controlled comparison would require additional generative benchmarks with low local content or MC benchmarks with comparable Sweden-specific factual content.

\subsection{Annotation Reliability}
\label{sec:discussion:annotation}
The Native-500 reference provides a stronger validation basis because every item has three independent annotations in its target language.
For explicit CS/CA, unanimous agreement ranges from 65--91\% across languages, with Gwet's AC1=.60--.91.
Danish illustrates the prevalence sensitivity of Fleiss'~$\kappa$: despite 84\% unanimous agreement, $\kappa$ is only .25 because 277/300 raw ratings are CS.
Norwegian Bokm{\aa}l is the weakest language-level explicit-CS/CA reliability case, with 65\% unanimous agreement, Fleiss'~$\kappa=.44$, Gwet's AC1=.60, and GPT-5.2 CS/CA accuracy of 86\%.
We therefore interpret NB-stratified content analyses with particular caution.
Overall, GPT-5.2 reaches 90.6\% CS/CA accuracy against the explicit three-annotator majority.
Four-way \texttt{required\_knowledge} is harder: 23/500 items have unresolved 1--1--1 human ties, and GPT-5.2 reaches 83.9\% accuracy on the 477 resolved items, or 80.0\% if all ties are counted as errors.
We therefore treat required-knowledge distinctions as noisier than the binary CS/CA variable.
Importantly, the main incremental-validity analysis uses total benchmark scores and is independent of individual annotation labels.
Annotation uncertainty affects the stratified analyses and descriptive content characterization rather than the core regression result.

\subsection{Generalizability}
\label{sec:discussion:generalizability}
Our evidence is restricted to five Nordic and Baltic target languages and DeepL-based MMLU proxies.
The exact $\tau_b$, $\Delta R^2$, and coefficient magnitudes should not be assumed to transfer one-to-one to human or LLM translations, because translation quality can change item difficulty and model rankings.
At the same time, translation cannot introduce source content that is absent from MMLU.
The observed content-overlap pattern therefore motivates a broader hypothesis: translated benchmarks should be most useful when the source benchmark covers knowledge similar to the native evaluation target, and less reliable when the native target is strongly local.
The language sample spans Germanic (SV, DA, NB), Finnic (ET), and Baltic (LT) languages, but remains geographically and typologically narrow.
Whether the proxy-validity gradient generalizes to typologically distant and lower-resource languages remains an open question.
Native benchmarks in other language settings can support such tests in future work.

\section{Conclusion}
\label{sec:conclusion}
We framed the question of whether translated benchmarks can stand in for native ones as a proxy-validity problem and tested whether translated MMLU scores carry incremental predictive value for native benchmark performance after controlling for model size and target-language proficiency.
Across 29 models and seven pairings in five Nordic and Baltic languages, we find a clear gradient rather than a binary answer.
Proxy validity is strongest when native benchmarks share broad subject or curriculum coverage with MMLU, as in Skolprov ($\Delta R^2=.128$, 95\% CI $[.038,.253]$).
It remains robust for some culturally loaded benchmarks such as INCLUDE-LT and NRK-Quiz, becomes marginal for DanishCitizenTest and INCLUDE-ET, and is null in the main specification for SwedishFacts.
A sensitivity analysis controlling for English MMLU performance supports the same overall pattern, indicating that general MMLU-content familiarity alone does not explain the observed proxy-validity differences.
These results offer practical guidance: translated benchmarks are most defensible when their source content aligns with the native evaluation target, but locally grounded competences may require native benchmarks rather than translated substitutes.

\section*{Limitations}
Our regression analyses cover 29 models, which is substantial for an evaluation study but still small for regression.
The effect sizes and bootstrap intervals should therefore be interpreted as uncertainty over the evaluated model sample rather than over a population of all LLMs.
Similarly, LOFO results are used as a conservative stability check.
The nominal $p<.05$ criterion across 17 refits is not multiplicity-corrected.

All translated proxies in this study use DeepL.
Other translation approaches, including human translation or LLM-based translation, may change absolute scores, rankings, and effect magnitudes.
A multi-translator design would be needed to separate content effects from translation-quality effects more directly.

SwedishFacts is the only generative benchmark in the study, so locality/content and response format remain entangled for this pairing.
The descriptive MC-only trend is useful for interpretation, but it is not a controlled format comparison.

The human reference covers 500 items, with 100 per language.
Although three annotations per item support multilingual IAA, 23 required-knowledge items have unresolved 1--1--1 ties, showing that fine-grained cultural versus regional distinctions can be ambiguous.

Benchmark-level CS proportions used in Figure~\ref{fig:proxy-gradient} are estimated by the automatic annotation pipeline rather than directly human-labeled.
On Native-500, CA recall is 80.7\%, with 35 CA$\to$CS versus 12 CS$\to$CA errors, suggesting that estimated CS shares may be slightly upward biased.
This may shift points horizontally in the descriptive gradient but does not affect the total-score regression outcomes.

Finally, the study covers five European target languages.
Evaluation on typologically distant and lower-resource languages is needed before treating the observed content/locality gradient as broadly language-general.

\section*{Ethics Statement}

This study involves human annotation of cultural-sensitivity and required-knowledge labels for benchmark items.
Professional native-language annotators were recruited through ProZ.com and compensated on individually agreed terms.

The final Native-500 reference contains three independent annotations per item in each language, corresponding to 15 pseudonymous annotator IDs in total, and no personally identifying information is reported.
Annotators were informed about the purpose of the study and intended use of their annotations.

GPT-5.2 is used for automatic annotation and for a separate model-assisted QA workflow.
To avoid using model-assisted corrections as the human gold standard, the IAA analyses and Native-500 meta-evaluation use reconstructed raw/pre-QA human submissions and majority consensus, before GPT-assisted consistency review.
Model-generated cultural-sensitivity labels may themselves reflect biases in the model's training data.
We therefore report multilingual human agreement, confusion patterns, and tie rates transparently (\S\ref{sec:annotation:metaeval}; Appendix~\ref{sec:app:annotation}).

All benchmark datasets used in this study are publicly available under their respective licenses.
We release our annotations, the annotation pipeline, and the analysis code to support reproducibility.

\section*{Acknowledgments}
\label{sec:ack}%
\makeatletter%
\ifacl@anonymize%
Acknowledgements removed for anonymization purposes.
\else%
The authors acknowledge the financial support by the Federal Ministry of Research, Technology and Space of Germany and by Sächsisches Staatsministerium für Wissenschaft, Kultur und Tourismus in the programme Center of Excellence for AI-research ``Center for Scalable Data Analytics and Artificial Intelligence Dresden/Leipzig'', project identification number: ScaDS.AI.
The authors gratefully acknowledge the computing time made available to them on the high-performance computer at the NHR Center of TU Dresden. This center is jointly supported by the Federal Ministry of Research, Technology and Space of Germany and the state governments participating in the NHR (www.nhr-verein.de/unsere-partner).
Evaluations were performed with computing resources granted by WestAI under project \texttt{westai0018} and by ZIH at TU Dresden for infrastructure used in automated evaluation computations.
\fi%
\makeatother

\bibliography{bibliography}

\appendix

\clearpage
\section{Extended Related Work}
\label{sec:app:related}

This appendix provides extended discussion of the related work summarized in \S\ref{sec:rel_work}, with more detailed descriptions of individual studies.

\subsection{Translation in Multilingual Evaluation}

\citet{Rajaee_2025} translate four NLP tasks: XNLI, PAWS-X (paraphrase identification), XCOPA (causal commonsense reasoning) and XStoryCloze (story completion) into 198 languages via NLLB.
They find that MT-based evaluations broadly track human-translated evaluations, although the strength of agreement varies across tasks and model settings.
They further show that commonly benchmarked low-resource languages can lead to systematic overestimation of model performance relative to a broader set of low-resource languages.

\citet{Vroe_2025} compare English, Finnish-HT, and Finnish-MT versions of ARC Challenge and observe a nuanced picture: 
while model-size rankings are broadly preserved, base and instruction-tuned models swap places depending on the translation variant, suggesting that translation effects interact with model type.

\citet{Agrawal_2024} provide a fine-grained analysis showing that translation inconsistencies in XNLI disproportionately impact low-resource languages, with accuracy gaps reaching 10 points.
At scale, \citet{Huang_2025} implement an MT$\rightarrow$post-editing pipeline across 16 languages and 10 tasks, demonstrating that post-editing can mitigate MT artifacts.

\citet{Artetxe_2023} demonstrate that carefully adapted translate-test can be competitive with cross-lingual transfer for classification, though the evaluation remains within translated English-origin content.

\citet{Ahuja_2023} evaluate generative LLMs across a wide set of NLP tasks in multiple languages (MEGA) and find that performance gaps are strongly language-dependent, with some languages showing near-English performance while others lag far behind.

\citet{Lai_2023} evaluate ChatGPT on non-English benchmarks and document substantial cross-lingual performance variation, with large gaps for some languages with low CommonCrawl representation.

\subsection{Native versus Translated Benchmarks}

\citet{Wu_2025} conduct a meta-analysis of over 2,000 multilingual benchmarks and report that localized benchmarks correlate more strongly with human preferences than translated ones.
However, the analysis does not control for general model strength, leaving open the possibility that the correlation difference is partly driven by the ``strong models are strong everywhere'' confound.

\citet{Singh_2025} build Global~MMLU with improved translations and cultural-sensitivity annotations (CS/CA labels) across 42 languages.
They demonstrate that model rankings shift when evaluation is restricted to culturally sensitive items, a key finding that motivates our own CS/CA-stratified analysis, but they do not compare against independently created native exams.

\citet{Keleg_2023} introduce DLAMA, a culturally diverse probing benchmark, and show that LLMs underperform on factual knowledge associated with non-Western cultures, providing direct evidence that knowledge gaps in LLMs are culturally patterned rather than uniform.

Closest to our work, \citet{Hupkes_2025_MultiLoKo} introduce MultiLoKo with locally sourced questions for 31 languages alongside human and machine translations, and quantify a substantial Locality Effect.
The reported performance gap between locally sourced and translated data exceeds 20 accuracy points for some model-language combinations.
However, MultiLoKo focuses on benchmark construction and descriptive comparisons.
It does not formalize proxy validity with statistical controls or attempt to disentangle the factors driving the observed gaps.
Our work extends this line by providing such controls and by analyzing whether proxy validity differs between culturally agnostic and culturally sensitive subsets.

\subsection{Translationese and Translation Artifacts}

\citet{Zhang_2019} demonstrate that translated MT test sets inflate human evaluation scores for machine translation systems and can alter system rankings, because evaluators find translated references more predictable.

\citet{Wang_2023} show that translationese in cross-lingual summarization data creates systematic discrepancies between automatic metrics and human evaluation, calling into question the reliability of evaluation on translated data.

\citet{Park_2024} provide evidence that models trained on translated data adapt to translation artifacts rather than learning genuine cross-lingual representations, inflating their apparent cross-lingual performance on benchmarks that share these artifacts.

\citet{Plazaortiz_2025} conduct an item-level analysis of MMLU translated into Spanish and confirm that a significant share of model errors trace back to translation problems including ambiguous answer choices, mistranslated domain terminology, and loss of pragmatic cues rather than to genuine knowledge gaps.

These findings collectively establish that translation effects are not mere noise but systematic biases that can fundamentally alter evaluation conclusions.

\subsection{Validity Frameworks for LLM Evaluation}

\citet{Salaudeen_2025} propose a validity-centered framework for AI evaluation that emphasizes claim-first reasoning and distinguishes criterion validity from proxy validity.
Their use of ``proxy validity'' refers to whether a benchmark score (observable) validly represents a latent construct (e.g., reasoning ability).
Our use of the term is narrower and more practical.
We ask whether one observable benchmark (translated) can substitute for another observable benchmark (native) in producing equivalent evaluation conclusions.

\citet{Bean_2025} conduct a systematic review of 445 LLM benchmarks from leading ML and NLP conferences and find widespread threats to construct validity, including vague phenomenon definitions, insufficient sampling, and missing reliability assessments, concerns that apply directly to translated benchmarks used as stand-ins for native tests.

\citet{Alaa_2025} focus specifically on the medical domain, drawing parallels between LLM benchmarks and psychological tests.
They demonstrate empirically that high scores on medical exam benchmarks (e.g., MedQA) do not predict performance on real clinical records, exposing a fundamental gap in predictive validity that resonates with our concern about translated benchmarks as proxies.

More broadly, \citet{Wallach_2025} argue that evaluating GenAI systems is fundamentally a social science measurement challenge and present a four-level framework grounded in measurement theory that highlights how current benchmarking practices often skip the crucial systematization step.

\citet{Weidinger_2025} call for maturing an evaluation science for generative AI, drawing lessons from transportation, aerospace, and pharmaceutical engineering where evaluation metrics must be applicable to real-world performance and iteratively refined.

On the empirical side, \citet{Burnell_2023} apply factor analysis to a large battery of LLM evaluations and find that capabilities are not monolithic but explained by multiple latent factors (reasoning, comprehension, language modeling).

Complementing this, \citet{Ilic_2024_AGA} analyze 591 LLMs and identify a strong general ability factor analogous to the $g$-factor in psychometrics.
Both findings motivate our use of statistical controls: 
without accounting for general model strength, any two benchmarks will correlate simply because strong models tend to perform well everywhere.

\citet{Kim_2025_BenchmarkProfiling} further show through benchmark profiling that individual benchmarks typically measure mixtures of abilities rather than single constructs, reinforcing that a translated benchmark may tap into a different ability mixture than a native exam even when both nominally test ``knowledge in language~$L$.''

\clearpage
\section{Appendix: Datasets}
\label{sec:app:datasets}

\subsection{G-MMLUX Construction}
\label{sec:app:datasets:gmmlux}

G-MMLUX is a unified subset that maps the culturally annotated items from GlobalMMLU \citep{Singh_2025} to their DeepL translations in each target language, including the English originals.
Before filtering, G-MMLUX contains 16{,}996 items (EN/DA/ET/LT/SV\,=\,2{,}850 each; NB\,=\,2{,}746).
We apply agreement filtering to exclude items with low annotator consensus in GlobalMMLU, removing 50 items per language for EN/DA/ET/LT/SV and 45 for NB, yielding 16{,}701 items in the final G-MMLUX (EN/DA/ET/LT/SV\,=\,2{,}800; NB\,=\,2{,}701).
All downstream analyses use this filtered version.

GlobalMMLU-Lite consists of 391 items after label aggregation (174 dev + 217 test); we use the 217-item test split in all evaluations.
DanishCitizenTest was deduplicated from the original 720 items to 668.

\subsection{Subject and Category Mapping}
\label{sec:app:datasets:mapping}

This appendix provides the complete subject-to-MMLU mapping, native subject translations, dataset statistics, and cross-validation details referenced in \S\ref{sec:datasets}.

\paragraph{Mapping confidence levels.}
To capture the ambiguity inherent in cross-benchmark subject alignment, we assigned each mapping a \emph{confidence level}: \textbf{direct} (1:1 MMLU subject match), \textbf{approximate} (plausible but not exact), or \textbf{no-equivalent} (category-level only).
Table~\ref{tab:confidence-compact} shows the distribution across datasets.

Notably, all native exam items fall into the \emph{approximate} or \emph{no-equivalent} categories, illustrating why subject-level comparison with MMLU is inherently imprecise.

\begin{table}[htbp]
\centering
\small
\begin{tabular}{@{}lrrr@{}}
\toprule
\textbf{Dataset} & \textbf{Direct} & \textbf{Approx.} & \textbf{No-equiv.} \\
\midrule
GlobalMMLU          & 2{,}850 & ---   & ---   \\
GlobalMMLU-Lite     & 217     & ---   & ---   \\
\midrule
Skolprov             & ---     & 465   & 80    \\
Exam-ET              & ---     & 1{,}130 & 484 \\
DanishCitizenTest    & ---     & 668   & ---   \\
\midrule
SwedishFacts         & ---     & ---   & 1{,}292 \\
INCLUDE-base (LT)    & ---     & ---   & 534   \\
INCLUDE-base (ET)    & ---     & ---   & 224   \\
NRK-Quiz-QA          & ---     & ---   & 3{,}600 \\
\bottomrule
\end{tabular}
\caption{
Items per mapping confidence level. 
\emph{Direct}: 1:1 MMLU subject match. 
\emph{Approx.}: plausible subject mapping with granularity or scope mismatch. 
\emph{No-equiv.}: no MMLU subject equivalent; mapped at category level only. 
Datasets without native subject labels (bottom block) are classified entirely via item-level annotation.
}
\label{tab:confidence-compact}
\end{table}

\paragraph{Mapping challenges: detailed examples.}
Three types of mismatches complicate direct subject-level alignment between native benchmarks and MMLU.

\emph{Granularity mismatches} arise when a native exam covers multiple difficulty levels that MMLU separates.
Estonian \textit{matemaatika} in Exam-ET spans content from basic algebra to university-level calculus, which MMLU distributes across \texttt{elementary\_mathematics}, \texttt{high\_school\_mathematics}, and \texttt{college\_mathematics}.
Similarly, \textit{füüsika} maps to \texttt{high\_school\_physics}, \texttt{college\_physics}, and \texttt{conceptual\_physics}.
Without item-level difficulty annotation, this ambiguity cannot be fully resolved.

\emph{Scope mismatches} occur when a single native subject conflates content from distinct MMLU subjects.
The Swedish \textit{matematik\_och\_fysikprovet} combines mathematics and physics in a single test, spanning three MMLU subjects.
Estonian \textit{ühiskonnaõpetus} (Social Studies) covers both government/politics and sociology.
The Danish citizenship and residence tests similarly span civics, history, and cultural knowledge that MMLU distributes across multiple subjects.

\begin{table}[htbp]
\centering
\small
\setlength{\tabcolsep}{0.4em}
\begin{tabular}{@{}p{0.48\columnwidth}p{0.44\columnwidth}@{}}
\toprule
\textbf{Native Subject} & \textbf{English} \\
\midrule
\multicolumn{2}{@{}l}{\textit{Skolprov (SV)}} \\[2pt]
kunskapsprov\_läkare       & Prof.\ Knowl.: Physician \\
kunskapsprov\_tandläkare   & Prof.\ Knowl.: Dentist \\
kunskapsprov\_audionom     & Prof.\ Knowl.: Audiologist \\
kunskapsprov\_apotekare    & Prof.\ Knowl.: Pharmacist \\
matematik\_och\_fysikprovet & Math.\ \& Physics Test \\
högskoleprovet             & Scholastic Apt.\ Test \\[3pt]
\multicolumn{2}{@{}l}{\textit{Exam-ET (ET)}} \\[2pt]
ajalugu                    & History \\
bioloogia                  & Biology \\
füüsika                    & Physics \\
geograafia                 & Geography \\
keemia                     & Chemistry \\
ühiskonnaõpetus             & Social Studies \\
eesti\_keel\_teise\_k.\_güm.  & Estonian as L2 (Gymn.) \\
eesti\_keel\_teise\_k.\_põhi. & Estonian as L2 (Basic) \\[3pt]
\multicolumn{2}{@{}l}{\textit{DanishCitizenTest (DA)}} \\[2pt]
citizenship                & Citizenship Test \\
residence                  & Residence Permit Test \\
\bottomrule
\end{tabular}
\caption{Native exam subject labels and English translations.}
\label{tab:app:subject-translations}
\end{table}

\emph{No-equivalent subjects} have no MMLU counterpart.
Language proficiency tests---such as \textit{eesti keel teise keelena} (Estonian as a Second Language) at gymnasium and basic school levels---assess L2 competence absent from MMLU.

\begin{table}[htbp]
\centering
\small
\setlength{\tabcolsep}{0.22em}
\renewcommand{\arraystretch}{1.05}
\begin{tabular}{@{}lcrrrrr@{}}
\toprule
\textbf{Dataset} & \textbf{L} & \textbf{Items} & \textbf{Subj.} & \textbf{M-subj.} & \textbf{M-cat.} & \textbf{Ann.} \\
\midrule
GlobalMMLU  & EN & 2{,}850 & 57 & 57 & 6 & --- \\
G-MMLU-Lite  & EN & 217     & 43 & 43 & 6 & 6 \\
\midrule
Skolprov     & SV & 545     & 6  & 9  & 3 & 5 \\
Exam-ET      & ET & 1{,}614 & 8  & 13 & 3 & 5 \\
DanishCT     & DA & 668     & 2  & 2  & 1 & 6 \\
\midrule
SwedishFacts & SV & 1{,}292 & --- & --- & --- & 6 \\
INCLUDE-LT   & LT & 534     & --- & --- & --- & 6 \\
INCLUDE-ET   & ET & 224     & --- & --- & --- & 5 \\
NRK-Quiz-QA  & NB & 3{,}600 & --- & --- & --- & 6 \\
\bottomrule
\end{tabular}
\caption{
Dataset overview.
\textbf{Subj.}: number of native subject labels.
\textbf{M-subj.}: number of distinct MMLU subjects covered by the mapping.
\textbf{M-cat.}: number of MMLU categories covered.
\textbf{Ann.}: number of distinct \texttt{subject\_category} labels assigned by GPT-5.2 annotation.
Datasets without native subject labels rely on item-level annotation for category assignment.
}
\label{tab:app:dataset-overview}
\end{table}

Vocabulary comprehension (\textsc{ord}) and reading comprehension (\textsc{läs}) in the Swedish Högskoleprov test linguistic aptitude rather than domain knowledge.

Professional licensing exams in Skolprov---\textit{kunskapsprov läkare} (physician), \textit{kunskapsprov tandläkare} (dentist), \textit{kunskapsprov audionom} (audiologist), and \textit{kunskapsprov apotekare} (pharmacist)---test specialized clinical knowledge that MMLU approximates at best through \texttt{professional\_medicine} or \texttt{clinical\_knowledge}, but without the vocational specificity of the Swedish exams.

These mismatches motivate our three-level confidence system (\emph{direct}, \emph{approximate}, \emph{no-equivalent}) and our recommendation that category-level rather than subject-level comparison is more appropriate for cross-benchmark analysis.

For detailed numbers, please refer to Table~\ref{tab:app:subject-mapping}.

\begin{table*}[htbp]
\centering
\small
\begin{tabular}{@{}p{2.8cm}p{3.0cm}p{4.5cm}lll@{}}
\toprule
\textbf{Dataset} & \textbf{Native Subject} & \textbf{MMLU Subject(s)} & \textbf{Cat.} & \textbf{Conf.} & \textbf{$n$} \\
\midrule
\multicolumn{6}{@{}l}{\textit{Skolprov (SV)}} \\
\midrule
 & kunskapsprov\_läkare & professional\_medicine, clinical\_knowledge, college\_medicine & MED & appr. & 132 \\
 & matematik\_och\_fysikprovet & high\_school\_mathematics, high\_school\_physics, college\_physics & STEM & appr. & 125 \\
 & kunskapsprov\_audionom & clinical\_knowledge, anatomy & MED & appr. & 85 \\
 & högskoleprovet & miscellaneous & OTH & no-eq. & 80 \\
 & kunskapsprov\_tandläkare & professional\_medicine, clinical\_knowledge, anatomy & MED & appr. & 70 \\
 & kunskapsprov\_apotekare & clinical\_knowledge, nutrition & MED & appr. & 53 \\
\midrule
\multicolumn{6}{@{}l}{\textit{Exam-ET (ET)}} \\
\midrule
 & ühiskonnaõpetus & hs\_gov\_and\_politics, sociology & SOC & appr. & 310 \\
 & ajalugu & hs\_european\_history, hs\_world\_history & HUM & appr. & 261 \\
 & eesti\_keel\_teise\_keelena\_güm. & miscellaneous & HUM & no-eq. & 246 \\
 & eesti\_keel\_teise\_keelena\_põhi. & miscellaneous & HUM & no-eq. & 238 \\
 & bioloogia & high\_school\_biology, college\_biology & STEM & appr. & 207 \\
 & keemia & high\_school\_chemistry, college\_chemistry & STEM & appr. & 133 \\
 & geograafia & high\_school\_geography & SOC & appr. & 111 \\
 & füüsika & hs\_physics, college\_physics, conceptual\_physics & STEM & appr. & 108 \\
\midrule
\multicolumn{6}{@{}l}{\textit{DanishCitizenTest (DA)}} \\
\midrule
 & citizenship & hs\_gov\_and\_politics, sociology & SOC & appr. & 336 \\
 & residence & hs\_gov\_and\_politics, sociology & SOC & appr. & 332 \\
\midrule
\multicolumn{6}{@{}l}{\textit{Datasets without native subjects}} \\
\midrule
SwedishFacts (SV) & \multicolumn{2}{l}{\textit{No native subjects; category via item-level annotation}} & mixed & no-eq. & 1{,}292 \\
INCLUDE-base (LT) & & & mixed & no-eq. & 534 \\
INCLUDE-base (ET) & & & mixed & no-eq. & 224 \\
NRK-Quiz-QA (NB)  & & & mixed & no-eq. & 3{,}600 \\
\bottomrule
\end{tabular}
\caption{
Native subject to MMLU mapping. 
\textbf{Cat.}: mapped MMLU category (STEM, HUM=Humanities, SOC=Social Sciences, MED=Medical, BUS=Business, OTH=Other). 
\textbf{Conf.}: mapping confidence---\emph{appr.}=approximate, \emph{no-eq.}=no equivalent. 
GlobalMMLU and GlobalMMLU-Lite are omitted as their subjects map directly (1:1) to MMLU. \textbf{$n$}: number of items.
}
\label{tab:app:subject-mapping}
\end{table*}

\begin{table*}[htbp]
\centering
\small
\setlength{\tabcolsep}{0.45em}
\begin{tabular}{@{}llrrrrll@{}}
\toprule
\textbf{Dataset} & \textbf{Lang.} & \textbf{Items} & \textbf{CA} & \textbf{CS} & \multicolumn{2}{c}{\textbf{Required Knowledge}} & \textbf{Time-sens.} \\
\midrule
GlobalMMLU       & EN & 2{,}850 & 2{,}058 & 792   & \multicolumn{2}{l}{none=9505; regional=1809; cultural=1253; dialect=84} & Yes=428 \\
GlobalMMLU-Lite  & EN & 217     & 121     & 96    & \multicolumn{2}{l}{none=134; cultural=42; regional=40; dialect=1}       & Yes=32 \\
\midrule
Skolprov          & SV & 545     & 467     & 78    & \multicolumn{2}{l}{none=529; regional=14; dialect=2}                    & Yes=66 \\
Exam-ET           & ET & 1{,}614 & 886     & 728   & \multicolumn{2}{l}{none=1120; regional=485; cultural=9}                 & Yes=229 \\
DanishCitizenTest & DA & 668     & 31      & 637   & \multicolumn{2}{l}{regional=558; cultural=77; none=33}                  & Yes=127 \\
\midrule
SwedishFacts      & SV & 1{,}292 & 28      & 1{,}264 & \multicolumn{2}{l}{regional=641; cultural=622; none=27; dialect=2}    & Yes=185 \\
INCLUDE-base (LT) & LT & 534     & 164     & 370   & \multicolumn{2}{l}{regional=294; none=187; cultural=53}                 & Yes=119 \\
INCLUDE-base (ET) & ET & 224     & 187     & 37    & \multicolumn{2}{l}{none=199; regional=20; dialect=3; cultural=2}        & Yes=4 \\
NRK-Quiz-QA       & NB & 3{,}600 & 512     & 3{,}088 & \multicolumn{2}{l}{cultural=1483; regional=1441; none=578; dialect=98} & Yes=1023 \\
\bottomrule
\end{tabular}
\caption{
Cultural annotation statistics. 
\textbf{CA}: culturally agnostic items. 
\textbf{CS}: culturally sensitive items. 
\textbf{Required Knowledge}: distribution of annotated knowledge types (note: GlobalMMLU counts reflect individual annotator votes, not aggregated labels). 
\textbf{Time-sens.}: items with time-sensitive content.
The annotation-derived counts are used only to characterize the benchmark collections. 
The annotation procedure, reliability analysis, and meta-evaluation are documented separately in Appendix~\ref{sec:app:annotation}.
}
\label{tab:app:cultural-annotations}
\end{table*}

\paragraph{Cross-validation with automated annotations.}
To validate category assignments, we compared them against the \texttt{subject\_category} labels produced by our GPT-5.2 annotation pipeline (\S\ref{sec:annotation}).
Native exam items use our manual category mappings; GlobalMMLU-Lite items use their direct MMLU categories.
Across 3{,}044 cross-validated items, agreement reaches \textbf{85.1\%} (Table~\ref{tab:app:cross-validation}).
The original MMLU taxonomy~\citep{hendrycks2021measuring} contains 57 subjects across four broad categories expanded to six by~\citet{Singh_2025}:
STEM (18), Humanities (13), Social Sciences (12), Medical (7), Business (5), and Other (2).

The 10 subjects with disagreements are attributable to inherent MMLU taxonomy ambiguities (law subjects classified as \textsc{humanities} vs.\ \textsc{social sciences}), heterogeneous catch-all categories, or low sample sizes ($n \leq 4$).
We found no clear cases where disagreements indicated a manual mapping error.
This supports category-level comparison as the more appropriate basis for cross-benchmark analysis; subject-level alignment is too noisy for reliable interpretation.

\begin{table*}[htbp]
\centering
\setlength{\tabcolsep}{0.15em}
\begin{tabular}{@{}llllrrl@{}}
\toprule
\textbf{Dataset} & \textbf{Native Subject} & \textbf{Our Cat.} & \textbf{GPT-5.2 Majority} & \textbf{$n$} & \textbf{Agree \%} & \textbf{Agreement} \\
\midrule
\multicolumn{7}{@{}l}{\textit{Native exam datasets (approximate / no-equivalent mappings)}} \\
\midrule
Skolprov & matematik\_och\_fysikprovet & STEM & STEM & 125 & 100.0 & \cmark \\
Skolprov & kunskapsprov\_läkare & MED & MED & 132 & 85.6 & \cmark \\
Skolprov & kunskapsprov\_tandläkare & MED & MED & 70 & 100.0 & \cmark \\
Skolprov & kunskapsprov\_audionom & MED & MED & 85 & 82.4 & \cmark \\
Skolprov & kunskapsprov\_apotekare & MED & MED & 53 & 71.7 & \cmark \\
Skolprov & högskoleprovet & OTH & HUM & 80 & 8.8 & \xmark \\
\midrule
Exam-ET & ühiskonnaõpetus & SOC & SOC & 310 & 100.0 & \cmark \\
Exam-ET & ajalugu & HUM & HUM & 261 & 95.0 & \cmark \\
Exam-ET & bioloogia & STEM & STEM & 207 & 98.1 & \cmark \\
Exam-ET & keemia & STEM & STEM & 133 & 99.2 & \cmark \\
Exam-ET & füüsika & STEM & STEM & 108 & 100.0 & \cmark \\
Exam-ET & geograafia & SOC & SOC & 111 & 57.7 & \cmark \\
Exam-ET & eesti\_keel\_teise\_keelena\_güm. & HUM & HUM & 246 & 93.1 & \cmark \\
Exam-ET & eesti\_keel\_teise\_keelena\_põhi. & HUM & HUM & 238 & 99.2 & \cmark \\
\midrule
DCT & citizenship & SOC & SOC & 336 & 70.5 & \cmark \\
DCT & residence & SOC & SOC & 332 & 76.2 & \cmark \\
\midrule
\multicolumn{7}{@{}l}{\textit{Selected MMLU-Lite subjects with disagreements (direct mappings)}} \\
\midrule
MMLU-Lite & international\_law & HUM & SOC & 5 & 0.0 & \xmark \\
MMLU-Lite & jurisprudence & HUM & SOC & 5 & 0.0 & \xmark \\
MMLU-Lite & professional\_law & HUM & SOC & 5 & 0.0 & \xmark \\
MMLU-Lite & human\_aging & MED & SOC & 10 & 10.0 & \xmark \\
MMLU-Lite & global\_facts & OTH & SOC & 9 & 0.0 & \xmark \\
MMLU-Lite & hs\_us\_history & HUM & SOC & 5 & 0.0 & \xmark \\
MMLU-Lite & miscellaneous & OTH & HUM & 29 & 27.6 & \xmark \\
\bottomrule
\end{tabular}
\caption{
Cross-validation of our category mapping against GPT-5.2 \texttt{subject\_category} annotations. \textbf{Our Cat.}: category assigned by our mapping. 
\textbf{GPT-5.2 Majority}: most frequent category in the independent annotation. 
\textbf{Agree \%}: proportion of items where our category matches the annotation. Overall agreement across 3{,}044 items: \textbf{85.1\%}. 
The bottom block shows disagreements originating from the MMLU taxonomy itself (law $\rightarrow$ humanities vs.\ social sciences), not from our mapping procedure. 
Category abbreviations: STEM, HUM=Humanities, SOC=Social Sciences, MED=Medical, BUS=Business, OTH=Other.
}
\label{tab:app:cross-validation}
\end{table*}

\subsection{Comparison Pairing Rationale}
\label{sec:app:datasets:comparison}

Each pairing in Table~\ref{tab:overview_comparison-pairings} matches a translated MMLU proxy with a native benchmark in the same language, designed to test proxy validity across a range of scenarios.

\paragraph{Curriculum-overlap pairings.}
\textbf{Exam-ET vs.\ MMLU\_ET}: Both are MC knowledge/exam QA. Exam-ET draws from the Estonian national curriculum across eight subjects; MMLU\_ET covers 57 subjects from a US-centric perspective. Our category mapping shows that 70\% of Exam-ET items (1{,}130 of 1{,}614) map at \emph{approximate} confidence, primarily in \textsc{stem} (biology, chemistry, physics), \textsc{humanities} (history), and \textsc{social sciences} (social studies, geography). The remaining 30\% (484 items) are language proficiency tests with no MMLU equivalent.

\textbf{Skolprov vs.\ MMLU\_SV}: Skolprov is rooted in Swedish educational standards and professional licensing, while MMLU\_SV is a DeepL translation of US-centric content. Of Skolprov's 545 items, 85\% (465) map at approximate confidence---primarily \textsc{medical} (340 items from the four professional licensing exams) and \textsc{stem} (125 items from the mathematics and physics test). The remaining 80 items from the Högskoleprov are a mixed aptitude test with no clean MMLU equivalent.

\paragraph{High-cultural-load pairings.}
\textbf{DanishCitizenTest vs.\ MMLU\_DA}: The Danish citizenship and residence permit tests specifically target knowledge about Danish society, culture, history, and democratic values---content that translated English-origin benchmarks are unlikely to capture well. With 95.4\% culturally sensitive items and the majority requiring regional knowledge (558 of 668 items annotated as \texttt{regional}), this pairing provides a strong test of whether translated proxies retain signal for locally anchored civic content.

\textbf{NRK-Quiz-QA vs.\ MMLU\_NB}: NRK-Quiz-QA covers Norwegian-specific general knowledge and cultural topics from public broadcasting quizzes, with 85.8\% culturally sensitive items. Unlike DanishCitizenTest (which tests civic/institutional knowledge), NRK-Quiz-QA spans broader cultural trivia including geography, language, and current affairs.

\paragraph{Middle-ground pairings.}
\textbf{INCLUDE (ET, LT) vs.\ MMLU\_ET/LT}: INCLUDE provides natively sourced exam questions from an independent multilingual benchmark effort, offering an additional native comparison point outside the national-exam collections. The Estonian subset (224 items) and Lithuanian subset (534 items) test region-specific academic and professional knowledge.

\textbf{SwedishFacts vs.\ MMLU\_SV}: SwedishFacts complements Skolprov by probing Sweden-specific factual knowledge in a generative (non-MC) format, with 97.8\% culturally sensitive items---the highest CS proportion in our study. Because it is the only non-MC benchmark in our study, we interpret it separately from the MC-based comparisons.
\clearpage
\section{Appendix: Annotation}
\label{sec:app:annotation}

\subsection{Annotation Schema and Interface}
\label{sec:app:annotation:guide}

This appendix provides the annotation schema, interface example, automatic-annotation prompt, human-reference reconstruction, inter-annotator agreement, and meta-evaluation details referenced in \S\ref{sec:annotation}.

\paragraph{Full annotation schema.}
Each item is a short benchmark item in one of five target languages.
For each item, annotators provide the following labels:

\begin{itemize}
\item \textbf{Required knowledge type (Q1).}
Whether answering correctly requires geographic, cultural, dialect-specific, or no specialized knowledge (exactly one of: \texttt{regional}, \texttt{cultural}, \texttt{dialect}, \texttt{none}).
The key criterion is \emph{necessity}: a knowledge type is selected only if it is required to answer correctly, not merely mentioned in passing.
A tie-break rule (dialect $>$ geographic $>$ cultural $>$ none) resolves cases where multiple types appear equally necessary.

\item \textbf{Time sensitivity (Q2).}
Whether the item contains explicit temporal or trend-based framing (e.g., ``currently'', ``in recent years'').
Annotators mark \texttt{Yes} only when such cues are explicitly present in the text.

\item \textbf{Trigger spans (Q3).}
Minimal, non-overlapping text spans that justify the Q1~decision, each labeled as \texttt{Geographic}, \texttt{Cultural}, or \texttt{Dialect}.
If the required knowledge is implicit, spans may be left empty without changing the Q1~decision.

\item \textbf{Conditional metadata (Q4--Q6).}
If Q1\,=\,\texttt{cultural}, annotators select one of seven culture groups (e.g., Western, Eastern Asian, Middle Eastern).
If Q1\,=\,\texttt{regional}, they select a world region and optionally a specific country in ISO~3166-1 format.
These fields are mutually exclusive by design to enforce schema consistency.

\item \textbf{Cultural sensitivity label (CS/CA).}
A binary label indicating whether the item is \emph{culturally sensitive}~(CS) or \emph{culturally agnostic}~(CA).
CA is assigned only when the item is free of any cultural, geographic, or dialect cues.
This label is not strictly determined by Q1: an item may be labeled~CS even if Q1\,=\,\texttt{none}, in borderline cases where cultural cues are present but not strictly necessary for answering.

\item \textbf{Subject category and explanation.}
One of six categories (STEM, Humanities, Social Sciences, Medical, Business, Other), based on the knowledge domain required to answer correctly, accompanied by a short English justification ($\leq$~250~characters).
\end{itemize}

\begin{figure*}[p]
    \centering
    \includegraphics[width=\textwidth,height=.86\textheight,keepaspectratio]{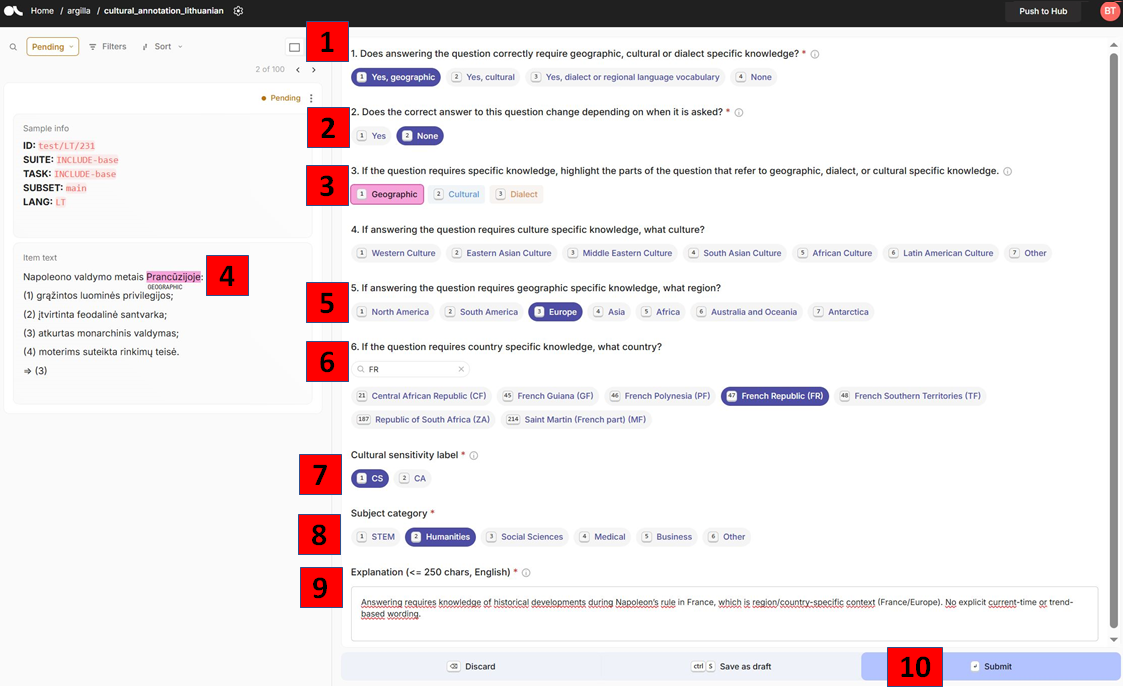}
    \caption{
    Argilla annotation interface shown through a completed Lithuanian item about Napoleonic rule in France.
    The interface includes the item text in the target language, guideline and navigation controls, required-knowledge selection, time-sensitivity flag, trigger-span highlighting, conditional culture/region/country metadata, explicit CS/CA label, subject category, explanation field, and submission controls.
    In this example, the annotator selected geographic/regional knowledge, marked no time sensitivity, highlighted the geographic trigger span ``Pranc\={u}zijoje'' (France), selected region \emph{Europe} and country \emph{France (FR)}, assigned the \emph{CS} label, chose \emph{Humanities} as the subject category, and provided a short explanation before submitting.
    }
    \label{fig:argilla-ui-3}
\end{figure*}

The annotation interface was deployed using Argilla\footnote{Argilla: \url{https://github.com/argilla-io/argilla}} hosted on Hugging Face Spaces and configured with the full schema described above.
Figure~\ref{fig:argilla-ui-3} shows a completed example, including the item panel, form fields, metadata fields, and submission controls.

\subsection{Annotation Prompt}
\label{sec:app:annotation:prompt}

Figure~\ref{fig:annotation-prompt} reproduces the system prompt used for automatic annotation.
The prompt asks GPT-5.2 to return the same schema as the human annotation interface in structured JSON format.

\begin{figure*}[p]
  \centering
  \begin{tcolorbox}[
    colback=gray!3,
    colframe=gray!50,
    fontupper=\scriptsize\ttfamily,
    title={\small\sffamily System Prompt for Cultural Sensitivity
      Annotation},
    left=4pt, right=4pt, top=4pt, bottom=4pt
  ]
You are an annotator for cultural sensitivity in exam-style
multiple-choice questions.

You will receive ONE item as plain text in language: \{LANG\}.

The text may include the answer key marker like "=> (1)". Do NOT use
the answer key to decide spans; you may ignore it.

\medskip
\textbf{Goal:}
Create a GlobalMMLU-style annotation that captures whether answering
correctly requires:
1) Cultural knowledge
2) Geographic (regional) knowledge
3) Dialect or regional language vocabulary knowledge
4) Temporal (time-sensitive) knowledge

\medskip
\textbf{Language requirement:}
All annotation fields and the explanation must be in English.
Only the "span" strings in references/other\_references must be
copied verbatim from the provided TEXT (and may be in \{LANG\}).

\medskip
\textbf{Definitions:}

\textit{Cultural knowledge:} Culture-specific facts, norms, values,
customs, religion, arts/literature, pop culture, historically rooted
traditions tied to a cultural group. Label as cultural if there is a
clear cultural anchor such as: religion or religious law/ritual (e.g.,
Sharia, Halakha, canon law) when relevant for answering; a named
canonical philosopher/scholar/text/school/tradition (esp.\ in
philosophy/psychology/political theory); culturally specific norms or
traditions. Treat named canonical figures/schools as a cultural anchor
(Western Culture), especially when the question asks about their
theories/views (e.g., Freud, Kant, Aristotle, Mill, Rawls, Bentham,
Hobbes, Locke, Marx, Weber, Durkheim). Do NOT treat "According to
Author \& Author (Year)" as cultural; this is domain-specific academic
recall. Also treat standard terminology from Western moral philosophy,
logic, and philosophy of mind (e.g., consequentialism, virtue ethics,
deontology, argument from ignorance, substance dualism) as a cultural
anchor with culture="Western Culture". Religion note: Do NOT map
religions mechanically to a region; set culture="Other" if no clear
region is anchored.

\textit{Geographic/regional knowledge:} Knowledge tied to specific
regions/countries. Important cues: Imperial/US customary units (inch,
mile, feet, Fahrenheit, pounds, mph) -> regional, region="North
America". Currency "\$" is ambiguous alone; only regional with
additional US anchor. US state names or US-specific events (e.g.,
World Series) -> regional. Named country-specific regulations (e.g.,
EMAS, GDPR, Sarbanes-Oxley) -> regional. Place-in-scenario: if a real
country/region is part of the scenario setup, label regional even if
the principle is general, UNLESS mentioned only as historical
background. Do NOT over-label: country mentions in historical passages
are not automatically regional.

\textit{Dialect knowledge:} Region/community-specific language
variations, slang, idioms, or dialect forms necessary to answer. Treat
common idioms (e.g., "Mom \& Pop", "blue-collar", "red tape", "y'all")
as dialect and include the idiom as a span.

\textit{Temporal knowledge:} Mark Yes if the item contains explicit time
references or trend/change language (e.g., "in recent years",
"currently", "has increased"), even if the fact might still be true.

\medskip
\textbf{Subject category} (always required): Assign exactly one of
STEM, Humanities, Social Sciences, Medical, Business, Other.

\medskip
\textbf{Decision rules:}
Only mark a knowledge type if it is necessary to answer correctly.
Choose exactly ONE required\_knowledge.
Tie-break: dialect > regional > cultural > none.
Set cultural\_sensitivity\_label="CA" ONLY if the item is clearly free
of cultural/geographic/dialect cues; otherwise "CS", even if
required\_knowledge="none" (borderline sensitivity).

\medskip
\textbf{Span rules:}
references must contain only spans justifying the chosen
required\_knowledge. Offsets are 0-indexed, end exclusive, matching
TEXT exactly. If knowledge is implicit, references can be [].

\medskip
\textbf{Output:} Return ONLY JSON:
\{"culture\_ann": \{"required\_knowledge": "none"|"cultural"|
"regional"|"dialect", "time\_sensitive": "Yes"|"No",
"cultural\_sensitivity\_label": "CS"|"CA", "subject\_category":
"STEM"|"Humanities"|"Social Sciences"|"Medical"|"Business"|"Other",
"culture": str|null, "region": str|null, "country": str|null,
"references": [\{"start": int, "end": int, "span": str,
"label": "Geographic"|"Cultural"|"Dialect"\}],
"other\_references": [...], "explanation": str (max 250 chars)\}\}
  \end{tcolorbox}
  \caption{
  Full system prompt for automatic cultural sensitivity annotation via GPT-5.2. 
  The placeholder \texttt{\{LANG\}} is replaced with the target language code. 
  All output fields are in English; only span extractions appear in the original language. 
  The prompt is also available in the accompanying code repository.
  }
  \label{fig:annotation-prompt}
\end{figure*}

\paragraph{API and reproducibility details.}
To support reproducibility, all automatic benchmark annotations were generated in February 2026 using the pinned model snapshot \texttt{gpt-5.2-2025-12-11} via the OpenAI API, with \texttt{reasoning.effort=high} and \texttt{max\_output\_tokens=8000}.
Each item was annotated in a single zero-shot call with the full system prompt (Figure~\ref{fig:annotation-prompt}).
The human-reference reconstruction and the separate model-assisted QA workflow are described below; importantly, the majority gold and IAA use raw/pre-QA human submissions rather than model-assisted corrections.

\subsection{Human Annotation Details}
\label{sec:app:annotation:details}

\subsubsection{Sampling and Recruitment}

For each target language (DA, ET, LT, NB, SV), we sampled 100 native-benchmark items.
Sampling was stratified using automatic metadata, including required knowledge, explicit CS/CA, time sensitivity, and subject category.
These automatic labels were used only for sampling and were removed before human annotation.
When a language contained multiple native datasets, we allocated the sample across datasets to preserve coverage.

Professional native-language annotators were recruited through ProZ.com.
The final Native-500 reference contains three independent annotations per item in each language, corresponding to 15 pseudonymous annotator IDs in total.
Annotators independently recorded \texttt{required\_knowledge}, explicit \texttt{cultural\_sensitivity\_label}, \texttt{time\_sensitive}, \texttt{subject\_category}, and the associated schema metadata.
Explicit CS/CA was a separate annotation field and was not mechanically derived from \texttt{required\_knowledge}.

\subsubsection{Raw/Pre-QA Reference Reconstruction}
\label{sec:app:annotation:reconstruction}

The annotation campaign produced multiple archived exports and workflow snapshots.
For the reliability and meta-evaluation analyses, we reconstruct the earliest usable \emph{raw/pre-QA} submission for each annotator and item.
We use these raw/pre-QA labels, rather than labels after GPT-assisted quality review, to avoid defining the human gold standard through the same model family used for automatic annotation.

The reconstruction starts from archived per-user raw snapshots, restores missing initial items from pre-validation ``finished'' snapshots where necessary, and uses the first stable completed but unvalidated submission when an earlier workflow record was incomplete.
Some archived records retain a workflow status such as \texttt{pending} despite populated annotation fields.
Inclusion is therefore based on annotation coverage, not the Argilla status flag.
The resulting reference contains complete values for all five evaluation fields on all 100 items for each of the 15 annotators.
The reconstruction reports and final item-level consensus files are retained with the accompanying analysis artifacts.

For each item, explicit CS/CA gold is the simple majority of the three independently annotated \texttt{cultural\_sensitivity\_label} values.
All 500 items have a unique explicit-CS/CA majority.
Required knowledge is aggregated independently: 477 items have a unique majority and 23 are unresolved 1--1--1 ties, which are excluded from the primary four-way meta-evaluation.
No model output is used to break these human ties.

\subsection{Inter-Annotator Agreement}
\label{sec:app:annotation:iaa}

We report IAA on the raw/pre-QA labels, before model-assisted review.
Confidence intervals are 95\% percentile intervals from 5{,}000 item-level bootstrap resamples (seed~20260501).
Because the explicit CS/CA label is the binary construct used in downstream stratification, it is the primary reliability result.

\begin{table}[t]
\centering
\setlength{\tabcolsep}{0.35em}
\renewcommand{\arraystretch}{1.05}

\begin{tabular}{@{}lcc@{}}
\toprule
\textbf{L} & \textbf{3/3 agree [95\% CI]} & \textbf{Pairwise [95\% CI]} \\
\midrule
DA & .84 [.77,.91] & .89 [.85,.94] \\
ET & .91 [.85,.96] & .94 [.90,.97] \\
LT & .80 [.72,.88] & .87 [.81,.92] \\
NB & .65 [.56,.74] & .77 [.71,.83] \\
SV & .83 [.76,.90] & .89 [.84,.93] \\
\bottomrule
\end{tabular}

\vspace{0.6em}

\begin{tabular}{@{}lcc@{}}
\toprule
\textbf{L} & \textbf{Fleiss $\kappa$ [95\% CI]} & \textbf{Gwet AC1 [95\% CI]} \\
\midrule
DA & .25 [$-.05$,.51] & .88 [.81,.93] \\
ET & .83 [.71,.92]   & .91 [.84,.96] \\
LT & .72 [.59,.82]   & .75 [.65,.85] \\
NB & .44 [.28,.57]   & .60 [.47,.72] \\
SV & .77 [.66,.86]   & .78 [.68,.87] \\
\bottomrule
\end{tabular}

\caption{
Raw/pre-QA inter-annotator agreement for the independently annotated explicit CS/CA field.
Each language has three annotators and 100 items.
Pairwise is mean pairwise agreement over the three annotator pairs.
Brackets show 95\% percentile bootstrap CIs from 5{,}000 item resamples.
}
\label{tab:iaa-csca-multilingual}
\end{table}

Table~\ref{tab:iaa-csca-multilingual} shows generally high explicit CS/CA agreement, but also illustrates why multiple agreement measures are useful.
Danish has the clearest prevalence effect: 277 of 300 raw ratings are CS.
Consequently, Fleiss'~$\kappa=.25$ has a wide CI crossing zero even though unanimous agreement is 84\%, mean pairwise agreement is .89, and prevalence-robust Gwet's AC1 is .88.

Required knowledge is more difficult than the binary CS/CA decision because it permits disagreement among four categories.
Table~\ref{tab:iaa-rk-multilingual} reports the corresponding consensus structure.

\begin{table}[t]
\centering
\setlength{\tabcolsep}{0.85em}
\renewcommand{\arraystretch}{1.05}

\begin{tabular}{@{}lrrrr@{}}
\toprule
\textbf{L} & \textbf{3/3} & \textbf{2/1} & \textbf{1/1/1} & \textbf{Unique} \\
\midrule
DA & 63 & 32 & 5  & 95  \\
ET & 81 & 19 & 0  & 100 \\
LT & 36 & 51 & 13 & 87  \\
NB & 37 & 59 & 4  & 96  \\
SV & 52 & 47 & 1  & 99  \\
\midrule
\textbf{All} & \textbf{269} & \textbf{208} & \textbf{23} & \textbf{477} \\
\bottomrule
\end{tabular}

\vspace{0.6em}

\begin{tabular}{@{}lc@{}}
\toprule
\textbf{L} & \textbf{Fleiss $\kappa$ [95\% CI]} \\
\midrule
DA & .20 [.06,.34] \\
ET & .67 [.58,.76] \\
LT & .33 [.23,.42] \\
NB & .37 [.26,.47] \\
SV & .48 [.38,.56] \\
\bottomrule
\end{tabular}

\caption{
Raw/pre-QA agreement structure for four-way \texttt{required\_knowledge}.
A 1/1/1 row is an unresolved three-way tie among the three annotators.
These 23 items are excluded from the primary four-way meta-evaluation.
}
\label{tab:iaa-rk-multilingual}
\end{table}

The largest ambiguity occurs in LT and NB, where unanimous agreement is 36\% and 37\%, respectively, and LT contributes 13 of the 23 unresolved ties.
This motivates reporting both the resolved-majority accuracy ($n=477$) and a conservative lower bound that counts all ties as errors.

\paragraph{Relationship to the model-assisted QA workflow.}
A later GPT-5.2-assisted review flagged schema consistency, missing fields, metadata constraints, and potential content-level issues for annotator review.
That workflow remains part of the annotation provenance and is useful for producing cleaned release annotations, but it is \emph{not} the source of the reported IAA or Native-500 majority gold.
This separation prevents the QA model from mechanically increasing its own measured agreement with the human reference.

\subsection{Meta-Evaluation Details}
\label{sec:app:annotation:metaeval}

This section provides the full split statistics, confusion matrices, construction checks, and per-language error breakdowns summarized in \S\ref{sec:annotation:metaeval}.

\subsubsection{GlobalMMLU-Lite Label Aggregation and Split}

GlobalMMLU-Lite provides annotations from 3--8 annotators per item (median:~4; 69.2\% of items have exactly~4 annotators).
We aggregate \texttt{required\_knowledge} via majority vote.
For ties, we distinguish two cases: 

(a)~simple ties between \textit{none} and exactly one non-none label are resolved in favor of the non-none label; 

(b)~complex ties involving multiple non-none labels are removed.

Of the 400~items, 144 (36.0\%) are unanimous, 185 (46.3\%) have a clear majority, 62 (15.5\%) are simple ties resolved via rule~(a), and 9 (2.3\%) are complex ties removed via rule~(b), yielding 391~items with resolved human labels.
The average majority share across all items is 77.0\%.

The provided \texttt{cultural\_sensitivity\_label} in GlobalMMLU follows a strict majority rule over the raw annotator votes: CA if strictly more than half of the annotators assigned \texttt{required\_knowledge}\,=\,\textit{none}, CS otherwise.
We verified this rule against the provided labels (100\% agreement on 391~items) and use these provided CS/CA labels directly in all GlobalMMLU-Lite evaluations.

From the 391~resolved items, we create a stratified split ($\sim$30\% dev, seed\,=\,13), yielding 174~dev and 217~test items.
The dev split is used for iterative prompt refinement; the test split is held out for final evaluation.
Table~\ref{tab:globalmmlu-split} reports the resulting label distribution.

\begin{table}[t]
  \centering
  \setlength{\tabcolsep}{0.42em}
  \renewcommand{\arraystretch}{1.05}
  \begin{tabular}{l r r r r r r r}
    \toprule
    \textbf{Split} & \textbf{n} & \rotatebox{60}{\textit{none}}
      & \rotatebox{60}{\textit{cult.}}
      & \rotatebox{60}{\textit{reg.}}
      & \rotatebox{60}{\textit{dial.}}
      & \textbf{CA} & \textbf{CS} \\
    \midrule
    Dev  & 174 & 103 & 15 & 55 & 1 &  90 &  84 \\
    Test & 217 & 123 & 33 & 59 & 2 & 110 & 107 \\
    \midrule
    Total & 391 & 226 & 48 & 114 & 3 & 200 & 191 \\
    \bottomrule
  \end{tabular}
  \caption{
  GlobalMMLU-Lite dev/test split after human label aggregation and tie resolution. 
  Columns show \texttt{required\_knowledge} counts after majority-vote aggregation and provided CS/CA labels.
  }
  \label{tab:globalmmlu-split}
\end{table}

\subsubsection{Prompt Calibration Error Patterns}
\label{sec:app:annotation:error-patterns}

Iterative refinement on the dev split identified five recurring misclassification patterns, each addressed through targeted prompt modifications:

(A)~under-detection of Western philosophical terminology as cultural anchors (e.g., consequentialism, virtue ethics);

(B)~under-detection of implicit US-centric regional cues such as imperial units, dollar references, and place-in-scenario setups;

(C)~under-detection of region-specific idioms as dialect cues (e.g., ``Mom \& Pop'', ``blue-collar'');

(D)~over-triggering of cultural labels on generic academic citations (``According to Author \& Author (Year)'');

and (E)~over-labeling of regional for incidental country mentions in historical passages.
All prompt modifications are motivated exclusively by dev-split findings.

The test split was never inspected during development.

\subsubsection{GlobalMMLU-Lite Test Evaluation}

Table~\ref{tab:conf-globalmmlu} shows the four-way \texttt{required\_knowledge} confusion matrix on the held-out GlobalMMLU-Lite test split.
The main errors are \textit{regional}$\to$\textit{none} and \textit{none}$\to$\textit{cultural}, reflecting the difficulty of detecting implicit regional and cultural cues in English-origin items.

\begin{table}[htbp]
  \centering
  \renewcommand{\arraystretch}{1.05}
  \begin{tabular}{l r r r r r}
    \toprule
    \textbf{H\,$\downarrow$\,\textbackslash\,G\,$\rightarrow$}
      & \textit{cult.} & \textit{dial.} & \textit{none} & \textit{reg.} & $\Sigma$ \\
    \midrule
    \textit{cultural}  & \textbf{25} &  0 &   8 &  0 &  33 \\
    \textit{dialect}   &  0 & \textbf{1} &   1 &  0 &   2 \\
    \textit{none}      & 14 &  0 & \textbf{102} &  7 & 123 \\
    \textit{regional}  &  3 &  0 &  23 & \textbf{33} &  59 \\
    \midrule
    $\Sigma$           & 42 &  1 & 134 & 40 & 217 \\
    \bottomrule
  \end{tabular}
  \caption{
  Required-knowledge confusion matrix on the GlobalMMLU-Lite test split ($n=217$). 
  H=human, G=GPT. Overall accuracy: 74.2\%.
  }
  \label{tab:conf-globalmmlu}
\end{table}

\subsubsection{Native-500 as External Meta-Evaluation}

The Native-500 sample was not used to tune the annotation prompt.
The item-level GPT annotations used for the native benchmarks were produced before the human-majority meta-evaluation and are joined to the reconstructed human reference only afterward.
Thus Native-500 is held out with respect to GlobalMMLU-Lite prompt calibration, while sharing the same annotation schema.

Tables~\ref{tab:conf-native-csca} and~\ref{tab:conf-native} report the two Native-500 confusion matrices.
The CS/CA matrix uses the explicit three-annotator majority for all 500 items.
The required-knowledge matrix uses the 477 items with a unique three-annotator majority.

\begin{table}[htbp]
\centering
\setlength{\tabcolsep}{0.82em}
\begin{tabular}{l rr r}
\toprule
\textbf{Human $\downarrow$ / GPT $\rightarrow$} & \textbf{CS} & \textbf{CA} & $\boldsymbol{\Sigma}$ \\
\midrule
CS & \textbf{307} & 12 & 319 \\
CA & 35 & \textbf{146} & 181 \\
\midrule
$\Sigma$ & 342 & 158 & 500 \\
\bottomrule
\end{tabular}
\caption{
Explicit CS/CA confusion matrix on Native-500. 
Human gold is the majority of the three independently annotated explicit CS/CA labels.
GPT uses its explicit \texttt{cultural\_sensitivity\_label}. 
Accuracy is 90.6\%; macro-F1 is 89.5\%.
}
\label{tab:conf-native-csca}
\end{table}

\begin{table}[htbp]
  \centering
  \setlength{\tabcolsep}{0.5em}
  \renewcommand{\arraystretch}{1.05}
  \begin{tabular}{l r r r r r}
    \toprule
    \textbf{H\,$\downarrow$\,\textbackslash\,G\,$\rightarrow$}
      & \textit{cult.} & \textit{dial.} & \textit{none} & \textit{reg.} & $\Sigma$ \\
    \midrule
    \textit{cultural}  & \textbf{66} & 1 & 3 & 44 & 114 \\
    \textit{dialect}   & 0 & \textbf{3} & 0 & 0 & 3 \\
    \textit{none}      & 10 & 0 & \textbf{166} & 7 & 183 \\
    \textit{regional}  & 8 & 0 & 4 & \textbf{165} & 177 \\
    \midrule
    $\Sigma$           & 84 & 4 & 173 & 216 & 477 \\
    \bottomrule
  \end{tabular}
  \caption{
  Required-knowledge confusion matrix on the 477 Native-500 items with a unique three-annotator majority. 
  Overall accuracy: 83.9\%; macro-F1: 82.4\%. 
  Counting all 23 unresolved human ties as errors gives a conservative lower bound of 400/500=80.0\%.
  }
  \label{tab:conf-native}
\end{table}

The dominant native four-way error is \textit{cultural}$\to$\textit{regional} (44 items), reflecting the fuzzy boundary between cultural and geographically anchored local knowledge.
Because explicit CS/CA is independently annotated, we do not collapse these four-way errors into the primary binary gold construction.

\subsubsection{CS/CA Construction Sensitivity}

The explicit human CS/CA field and a CS/CA label mechanically derived from majority \texttt{required\_knowledge} differ on 10/500 items (2\%).
This confirms that the two human fields are correlated but non-redundant.
Table~\ref{tab:csca-construction-sensitivity} therefore distinguishes the primary CS/CA evaluation from two sensitivity constructions.

\begin{table}[htbp]
\centering
\setlength{\tabcolsep}{0.35em}
\renewcommand{\arraystretch}{1.08}
\begin{tabular}{@{}p{0.40\columnwidth}cp{0.30\columnwidth}@{}}
\toprule
\textbf{Construction} & \textbf{Acc.} & \textbf{Role} \\
\midrule
Human explicit \newline vs.\ GPT explicit & 90.6\% & primary; downstream-aligned \\
\addlinespace[0.3em]
Human explicit \newline vs.\ GPT RK-derived & 92.8\% & sensitivity \\
\addlinespace[0.3em]
Human RK-derived \newline vs.\ GPT RK-derived & 94.4\% & legacy diagnostic \\
\bottomrule
\end{tabular}
\caption{
Native-500 CS/CA meta-evaluation under different constructions.
``RK-derived'' maps \texttt{none}$\to$CA and non-\texttt{none}$\to$CS.
The primary result compares the independently annotated explicit human field with GPT's explicit field.
}
\label{tab:csca-construction-sensitivity}
\end{table}

\clearpage
\subsubsection{Native Error Distribution by Language}

Table~\ref{tab:csca-by-lang} reports explicit-CS/CA mismatches by language.
Errors are asymmetric: the model more often predicts CS for human-CA items than the reverse, especially in Norwegian Bokm{\aa}l and Estonian.

\begin{table}[htbp]
  \centering
  \setlength{\tabcolsep}{0.45em}
  \renewcommand{\arraystretch}{1.05}
  \begin{tabular}{l c c c c}
    \toprule
    \textbf{Lang.} & \textbf{Acc.} & \textbf{Errors} & \textbf{CS$\to$CA} & \textbf{CA$\to$CS} \\
    \midrule
    DA & 92\% &  8 & 5 & 3 \\
    ET & 90\% & 10 & 0 & 10 \\
    LT & 92\% &  8 & 2 & 6 \\
    NB & 86\% & 14 & 2 & 12 \\
    SV & 93\% &  7 & 3 & 4 \\
    \bottomrule
  \end{tabular}
  \caption{
  Explicit CS/CA performance by language on Native-500. 
  CS$\to$CA denotes a human-CS/GPT-CA miss.
  CA$\to$CS denotes a human-CA/GPT-CS false alarm. 
  NB is the weakest language-level case (86\% accuracy; 12/14 errors are CA$\to$CS).
  }
  \label{tab:csca-by-lang}
\end{table}

Table~\ref{tab:rk-by-lang} reports four-way required-knowledge accuracy by language, using only items with a unique three-annotator majority.
The table also shows how many items per language are excluded as unresolved ties.

\begin{table}[htbp]
\centering
\setlength{\tabcolsep}{0.35em}
\begin{tabular}{l r r r r}
\toprule
\textbf{L} & \textbf{Resolved $n$} & \textbf{Correct} & \textbf{Accuracy} & \textbf{Ties} \\
\midrule
DA & 95 & 83 & 87.4\% & 5 \\
ET & 100 & 96 & 96.0\% & 0 \\
LT & 87 & 77 & 88.5\% & 13 \\
NB & 96 & 72 & 75.0\% & 4 \\
SV & 99 & 72 & 72.7\% & 1 \\
\midrule
\textbf{All} & 477 & 400 & 83.9\% & 23 \\
\bottomrule
\end{tabular}
\caption{
Four-way \texttt{required\_knowledge} accuracy by language using only items with a unique three-annotator majority.
}
\label{tab:rk-by-lang}
\end{table}

The required-knowledge results are strongest for Estonian and weakest for Swedish and Norwegian Bokm{\aa}l.
These differences support the interpretation in the main text that fine-grained knowledge-type labels are noisier than binary CS/CA labels.

\clearpage
\section{Proxy Validity: Supplementary Material}
\label{sec:app:proxy}

\subsection{Model Overview}
\label{sec:app:proxy:models}

Table~\ref{tab:app:models} lists all 29 models evaluated in our study, grouped by size category.

\begin{table}[htbp]
\centering
\setlength{\tabcolsep}{0.3em}
\begin{tabular}{@{}llrr@{}}
\toprule
\textbf{Cat.} & \textbf{Model} & \textbf{B} & \textbf{BA} \\
\midrule
\multicolumn{4}{@{}l}{\textit{M ($\geq$70B)}} \\[2pt]
 & c4ai-command-a-03-2025-111B  & 111 & 111 \\
 & Apertus-70B-Instruct-2509    & 70  & 70  \\
 & Llama-4-Scout-17B-16E        & 109 & 17  \\
 & Mistral-Large-2411            & 123 & 123 \\
 & Llama-3.1-70B-Instruct       & 70  & 70  \\
 & Qwen2.5-72B-Instruct         & 72  & 72  \\
 & Xwen-72B-Chat                & 72  & 72  \\[3pt]
\multicolumn{4}{@{}l}{\textit{S (15--32B)}} \\[2pt]
 & aya-expanse-32b              & 32  & 32  \\
 & Phi-4-15B-Instruct           & 15  & 15  \\
 & gpt-oss-20b                  & 20  & 3.6 \\
 & Nemotron-3-Nano-30B-A3B      & 30  & 3   \\
 & Qwen3-30B-A3B-2507           & 30  & 3   \\
 & EuroLLM-22B-Instruct-Prev.   & 22  & 22  \\
 & Mistral-Small-24B-2501       & 24  & 24  \\
 & gemma-3-27b-it               & 27  & 27  \\[3pt]
\multicolumn{4}{@{}l}{\textit{XS (7--9B)}} \\[2pt]
 & salamandra-7b-instruct       & 7   & 7   \\
 & Teuken-7B-instruct-v0.6      & 7   & 7   \\
 & Llama-3.1-8B-Instruct        & 8   & 8   \\
 & Qwen3-8B                     & 8   & 8   \\
 & Falcon3-7B-Instruct          & 7   & 7   \\
 & EuroLLM-9B-Instruct          & 9   & 9   \\
 & aya-expanse-8b               & 8   & 8   \\[3pt]
\multicolumn{4}{@{}l}{\textit{XXS ($\leq$1.7B)}} \\[2pt]
 & SmolLM2-360M-Instruct        & 0.4 & 0.4 \\
 & Qwen3-0.6B                   & 0.6 & 0.6 \\
 & LFM2-700M                    & 0.7 & 0.7 \\
 & gemma-3-1b-it                & 1   & 1   \\
 & Falcon3-1B-Instruct          & 1.7 & 1.7 \\
 & Llama-3.2-1B-Instruct        & 1.2 & 1.2 \\
 & bloomz-560m                  & 0.6 & 0.6 \\
\bottomrule
\end{tabular}
\caption{
Models evaluated in our study. 
\textbf{B}: total parameters (billions). 
\textbf{BA}: active parameters (billions; differs from B for mixture-of-experts models).
All evaluated checkpoints are instruction-tuned. 
Model labels are shortened, with exact identifiers in the released harness configuration. 
Regressions use B (total parameters). 
Inference uses NVIDIA H100 GPUs (94\,GiB HBM2e).
M-category models use 4-way tensor parallelism, others use a single GPU.
}

\label{tab:app:models}
\end{table}

\subsection{Data Foundation}
\label{sec:app:proxy:data}

Table~\ref{tab:app:data-foundation} summarizes the item-level data used for the ranking and incremental-validity analyses.
After merging benchmark outputs, CS/CA annotations, and per-model LM Evaluation Harness results, the final item-level analysis file contains 886{,}762 rows across 29 models.\footnote{The merged analysis data and data-processing pipeline are available at \GTDataProcessingRepo.}
This total decomposes as follows: G-MMLUX (16{,}701 items across 6~languages), native benchmarks excluding SwedishFacts (7{,}185 items), SwedishFacts (1{,}292 accuracy rows), and Belebele (5{,}400 items across 5~languages), giving $30{,}578 \times 29 = 886{,}762$ rows.
These balanced subsets are used only for the ranking analyses.
The main incremental-validity regressions use total benchmark scores rather than balanced CA/CS subsets.

\begin{table*}[htbp]
\centering
\begin{tabular}{@{}llrrrrrl@{}}
\toprule
\textbf{Group} & \textbf{Dataset} & \textbf{Lang.} & \textbf{Items} & \textbf{CA} & \textbf{CS} & \textbf{Balanced} & \textbf{Used in} \\
\midrule
\multicolumn{8}{@{}l}{\textit{Translated proxy benchmarks (MMLU)}} \\
\midrule
MMLU-only & G-MMLUX & DA & 2{,}800 & 2{,}059 & 741 & 741+741 & A, B \\
MMLU-only & G-MMLUX & EN & 2{,}800 & 2{,}059 & 741 & 741+741 & A \\
MMLU-only & G-MMLUX & ET & 2{,}800 & 2{,}059 & 741 & 741+741 & A, B \\
MMLU-only & G-MMLUX & LT & 2{,}800 & 2{,}059 & 741 & 741+741 & A, B \\
MMLU-only & G-MMLUX & NB & 2{,}701 & 2{,}008 & 693 & 693+693 & A, B \\
MMLU-only & G-MMLUX & SV & 2{,}800 & 2{,}059 & 741 & 741+741 & A, B \\
\midrule
\multicolumn{8}{@{}l}{\textit{Native benchmarks}} \\
\midrule
non-MMLU & DanishCitizenTest & DA & 668 & 31 & 637 & 31+31\textsuperscript{\dag} & B \\
non-MMLU & Exam-ET & ET & 1{,}614 & 886 & 728 & \multirow{2}{*}{765+765} & A, B \\
non-MMLU & INCLUDE-base & ET & 224 & 187 & 37 & & A, B \\
non-MMLU & INCLUDE-base & LT & 534 & 164 & 370 & 164+164 & A, B \\
non-MMLU & NRK-Quiz-QA & NB & 3{,}600 & 512 & 3{,}088 & 512+512 & A, B \\
non-MMLU & Skolprov & SV & 545 & 467 & 78 & \multirow{2}{*}{495+495} & A, B \\
non-MMLU & SwedishFacts & SV & 1{,}292 & 28 & 1{,}264 & & A, B \\
\midrule
\multicolumn{8}{@{}l}{\textit{Language proficiency}} \\
\midrule
--- & Belebele & 5 langs & 5{,}400 & \multicolumn{2}{c}{(not annotated)} & --- & Incr. \\
\bottomrule
\end{tabular}
\caption{
Data foundation for the proxy validity analysis. 
\textbf{Items}: unique annotated items. \textbf{CA}/\textbf{CS}: 
culturally agnostic/sensitive item counts. 
\textbf{Balanced}: 
items per class in the balanced CA/CS subsets used for Block~A ($\min(n_\text{CA}, n_\text{CS})$ per group $\times$ language). 
\textbf{Used in}: A\,=\,Block~A (CA vs.\ CS ranking), B\,=\,Block~B (native vs.\ proxy ranking), Incr.\,=\,incremental proxy validity (\S\ref{sec:proxy:incremental}). 
\textsuperscript{\dag}DA non-MMLU excluded from Block~A interpretation due to insufficient CA items ($n=31$). For ET and SV, multiple native datasets are pooled before balancing.
}
\label{tab:app:data-foundation}
\end{table*}

\paragraph{Evaluation infrastructure.}
All evaluations use our custom fork of the LM Evaluation Harness\footnote{LM Evaluation Harness: \url{https://github.com/EleutherAI/lm-evaluation-harness}; our fork with native benchmark task implementations: \GTHarnessRepo.} \citep{eval-harness}.
Inference runs on NVIDIA H100 GPUs (94\,GiB HBM2e).
Models in categories S, XS, and XXS use a single GPU, while M-category models use 4-way tensor parallelism.

\paragraph{Balanced CA/CS subsets.}
For the ranking analysis in \S\ref{sec:proxy:ranking}, we construct balanced CA/CS subsets: within each benchmark group (MMLU-only and non-MMLU native) and language, we sample $\min(n_\text{CA}, n_\text{CS})$ items per class to ensure that accuracy differences between CA and CS rankings are not driven by sample size imbalance.

\subsection{Block A: CA vs.\ CS Ranking Consistency (Details)}
\label{sec:app:proxy:blocka}

Table~\ref{tab:app:block-a} provides the full Block~A results.
The Danish native CA/CS comparison is shown for completeness but not interpreted in the main text because only 31 CA items are available.

\begin{table}[htbp]
\centering
\small
\setlength{\tabcolsep}{0.55em}
\begin{tabular}{@{}llrrccr@{}}
\toprule
\textbf{Group} & \textbf{L} & \textbf{$n_\text{CA}$} & \textbf{$n_\text{CS}$} & \textbf{$\tau_b$} & \textbf{[95\% CI]} & \textbf{$p$} \\
\midrule
\multicolumn{7}{@{}l}{\textit{MMLU-only}} \\[2pt]
 & DA & 741 & 741 & 0.91 & [.82,\,.97] & $<$10\textsuperscript{-11} \\
 & EN & 741 & 741 & 0.91 & [.85,\,.97] & $<$10\textsuperscript{-11} \\
 & ET & 741 & 741 & 0.91 & [.84,\,.97] & $<$10\textsuperscript{-11} \\
 & LT & 741 & 741 & 0.94 & [.86,\,.98] & $<$10\textsuperscript{-12} \\
 & NB & 693 & 693 & 0.91 & [.82,\,.97] & $<$10\textsuperscript{-11} \\
 & SV & 741 & 741 & 0.91 & [.82,\,.97] & $<$10\textsuperscript{-11} \\[3pt]
\multicolumn{7}{@{}l}{\textit{non-MMLU (native benchmarks)}} \\[2pt]
 & DA\textsuperscript{\dag} & 31 & 31 & 0.66 & [.48,\,.80] & $<$10\textsuperscript{-5} \\
 & ET & 765 & 765 & 0.87 & [.77,\,.94] & $<$10\textsuperscript{-10} \\
 & LT & 164 & 164 & 0.79 & [.66,\,.90] & $<$10\textsuperscript{-8} \\
 & NB & 512 & 512 & 0.88 & [.79,\,.96] & $<$10\textsuperscript{-10} \\
 & SV & 495 & 495 & 0.63 & [.43,\,.77] & $<$10\textsuperscript{-5} \\
\bottomrule
\end{tabular}
\caption{Block~A: CA vs.\ CS ranking consistency (Kendall's $\tau_b$) within MMLU-only and non-MMLU native benchmarks, using balanced CA/CS subsets. \textsuperscript{\dag}DA non-MMLU: excluded from main-text interpretation due to insufficient CA items ($n_\text{CA}=31$; ${\approx}$1 item per model).}
\label{tab:app:block-a}
\end{table}

\paragraph{Methodological note: balanced subsets.}
To ensure that CA vs.\ CS ranking differences are not confounded by unequal sample sizes, we construct balanced subsets by sampling $\min(n_\text{CA}, n_\text{CS})$ items per class within each benchmark group and language.
For MMLU-only, this yields 741 items per class per language, except for Norwegian Bokm{\aa}l, where 693 items per class are available because fewer translations are present.
For native benchmarks, the balancing is constrained by the minority class: ET and SV pool multiple datasets before balancing; DA is limited by only 31~CA items in DanishCitizenTest.
Sampling is reproducible via a fixed random seed ($s=42$).

\paragraph{Interpretation of the SV non-MMLU result ($\tau_b = 0.63$).}
The low Swedish consistency is a compositional effect rather than a substantive finding about cultural sensitivity.
Skolprov contributes 467 of 495 CA items (86\% CA, primarily STEM and medical), while SwedishFacts contributes 1{,}264 of 1{,}342 CS items (98\% CS, primarily cultural trivia).
The CA vs.\ CS comparison therefore effectively contrasts two distinct benchmarks with different knowledge requirements, rather than two facets of the same construct.
This is reflected in the Block~B results (\S\ref{sec:proxy:ranking}), where Skolprov alone achieves $\tau_b = 0.89$ against its MMLU proxy while SwedishFacts achieves only $\tau_b = 0.51$.

\subsection{Block B: Native vs.\ Proxy Ranking (Details)}
\label{sec:app:proxy:blockb}

Table~\ref{tab:app:block-b-full} extends Table~\ref{tab:proxy-rank-b} with the full set of statistics including item counts and top-3 overlap.

\begin{table*}[t]
\centering
\begin{tabular}{@{}llcrrcccccc@{}}
\toprule
\textbf{Native} & \textbf{L} & \textbf{Sub.} & $\boldsymbol{n}_\textbf{nat}$ & $\boldsymbol{n}_\textbf{prx}$ & \textbf{$\tau_b$} & \textbf{[95\% CI]} & $\boldsymbol{p}$ & \textbf{T3} & \textbf{T5} \\
\midrule
Skolprov       & SV & ALL & 545   & 2800 & .89 & [.78,\,.96] & $<$10\textsuperscript{-11} & 1/3 & 5/5 \\
               &    & CA  & 467   & 2059 & .88 & [.78,\,.95] & $<$10\textsuperscript{-10} & 2/3 & 4/5 \\
               &    & CS  & 78    & 741  & .77 & [.66,\,.86] & $<$10\textsuperscript{-8}  & 1/3 & 4/5 \\
\midrule
Exam-ET        & ET & ALL & 1614  & 2800 & .88 & [.78,\,.96] & $<$10\textsuperscript{-10} & 2/3 & 5/5 \\
               &    & CA  & 886   & 2059 & .83 & [.72,\,.91] & $<$10\textsuperscript{-9}  & 1/3 & 4/5 \\
               &    & CS  & 728   & 741  & .88 & [.78,\,.95] & $<$10\textsuperscript{-10} & 2/3 & 5/5 \\
\midrule
INCLUDE        & LT & ALL & 534   & 2800 & .87 & [.79,\,.93] & $<$10\textsuperscript{-10} & 1/3 & 5/5 \\
               &    & CA  & 164   & 2059 & .82 & [.73,\,.90] & $<$10\textsuperscript{-9}  & 2/3 & 5/5 \\
               &    & CS  & 370   & 741  & .85 & [.74,\,.93] & $<$10\textsuperscript{-10} & 2/3 & 5/5 \\
\midrule
NRK-Quiz       & NB & ALL & 3600  & 2701 & .85 & [.76,\,.92] & $<$10\textsuperscript{-10} & 1/3 & 4/5 \\
               &    & CA  & 512   & 2008 & .82 & [.73,\,.90] & $<$10\textsuperscript{-9}  & 1/3 & 3/5 \\
               &    & CS  & 3088  & 693  & .88 & [.79,\,.94] & $<$10\textsuperscript{-10} & 2/3 & 5/5 \\
\midrule
DanishCT       & DA & ALL & 668   & 2800 & .82 & [.73,\,.90] & $<$10\textsuperscript{-13} & 1/3 & 3/5 \\
               &    & CA  & 31    & 2059 & .66 & [.53,\,.79] & $<$10\textsuperscript{-5}  & 3/3 & 5/5 \\
               &    & CS  & 637   & 741  & .84 & [.75,\,.90] & $<$10\textsuperscript{-9}  & 1/3 & 4/5 \\
\midrule
INCLUDE        & ET & ALL & 224   & 2800 & .72 & [.58,\,.83] & $<$10\textsuperscript{-7}  & 2/3 & 3/5 \\
               &    & CA  & 187   & 2059 & .68 & [.54,\,.80] & $<$10\textsuperscript{-6}  & 1/3 & 3/5 \\
               &    & CS  & 37    & 741  & .72 & [.54,\,.85] & $<$10\textsuperscript{-7}  & 0/3 & 4/5 \\
\midrule
SwedishFacts   & SV & ALL & 1292  & 2800 & .51 & [.25,\,.74] & $<$10\textsuperscript{-3}  & 1/3 & 3/5 \\
               &    & CA  & 28    & 2059 & .50 & [.23,\,.71] & $<$10\textsuperscript{-3}  & 1/3 & 3/5 \\
               &    & CS  & 1264  & 741  & .52 & [.26,\,.74] & $<$10\textsuperscript{-4}  & 1/3 & 2/5 \\
\bottomrule
\end{tabular}
\caption{Block~B: full native-vs-proxy ranking consistency. \textbf{Sub.}: subset (ALL, CA-only, CS-only). $n_\text{nat}$/$n_\text{prx}$: number of items on native/proxy side. \textbf{T3}/\textbf{T5}: top-3/top-5 model overlap. The proxy is always G-MMLUX in the corresponding language. All comparisons use 29~models.}
\label{tab:app:block-b-full}
\end{table*}

\begin{table*}[t]
\centering
\setlength{\tabcolsep}{0.32em}
\begin{tabular}{@{}llrrrrccl@{}}
\toprule
\textbf{Native} & \textbf{L} & $\boldsymbol{R^2_\text{base}}$ & $\boldsymbol{R^2_\text{ext}}$ & $\boldsymbol{\Delta R^2}$ \textbf{[95\% CI]} & $\boldsymbol{\beta_\text{prx}}$ \textbf{[95\% CI]} & $\boldsymbol{p_F}$ & \textbf{LOFO} & \textbf{Class} \\
\midrule
Skolprov       & SV & .856 & .984 & .128 [.038,.253] & 1.056 [.798,1.187] & $<10^{-13}$ & 17/17 & robust \\
INCLUDE        & LT & .890 & .965 & .076 [.021,.148] & 0.938 [.667,1.126] & $<10^{-7}$ & 17/17 & robust \\
NRK-Quiz       & NB & .934 & .975 & .041 [.009,.087] & 0.463 [.289,.592] & $<10^{-6}$ & 17/17 & robust \\
Exam-ET        & ET & .954 & .979 & .024 [.010,.044] & 0.515 [.340,.852] & $<10^{-5}$ & 17/17 & robust \\
INCLUDE        & ET & .818 & .877 & .059 [.000,.199] & 0.842 [$-.021$,1.300] & .002 & 16/17 & marginal \\
DanishCT       & DA & .951 & .960 & .009 [.000,.027] & 0.309 [.006,.575] & .027 & 14/17 & marginal \\
SwedishFacts   & SV & .353 & .388 & .035 [.000,.163] & 0.140 [$-.096$,.329] & .242 & 0/17 & null \\
\bottomrule
\end{tabular}
\caption{
Full all-item incremental proxy-validity results with model-bootstrap uncertainty.
The baseline model predicts native performance from $\log(\text{total parameters})$ and Belebele.
The extended model adds the same-language translated MMLU proxy.
\textbf{LOFO} reports how many of the 17 leave-one-family-out refits retain positive $\beta_\text{proxy}$ with nominal $p<.05$.
The evidence class follows the fixed definition in the main text.
}
\label{tab:app:incremental-full}
\end{table*}

\paragraph{Note on the DanishCT CA result ($\tau_\text{CA} = 0.66$, T3\,=\,3/3, T5\,=\,5/5).}
The seemingly contradictory pattern---low $\tau_b$ but perfect top-$k$ overlap---arises because only 31~CA items are available in DanishCitizenTest, yielding noisy accuracy estimates.
The top models are sufficiently strong to maintain their relative order despite this noise, but mid-range and lower-ranked models swap positions frequently, pulling down the global correlation.

\paragraph{Methodology: Kendall's $\tau_b$.}
We use Kendall's $\tau_b$ (the variant correcting for ties) rather than Spearman's $\rho$ for two reasons:
(1)~with $n=29$ models, Kendall's $\tau_b$ is less sensitive to large rank displacements because it counts pairwise concordances rather than squared rank differences; and 
(2)~$\tau_b$ explicitly corrects for tied ranks, which are common when models achieve identical accuracy on small item sets.

Given $n=29$ and the presence of tied ranks, we interpret the reported $p$-values together with the bootstrap CIs rather than as standalone evidence.
Bootstrap CIs use 2{,}000 resamples over models; each resample draws 29 models with replacement and recomputes $\tau_b$.

\paragraph{Methodology: top-$k$ overlap.}
Top-$k$ overlap counts the number of models appearing in the top-$k$ positions of \emph{both} the native and proxy rankings (using dense ranks, where ties share the same rank).
We report $k \in \{3, 5\}$ as these correspond to the practical question of whether the ``best few'' models are the same regardless of which benchmark is used.

\subsection{CS/CA-Stratified Incremental Proxy Validity}
\label{sec:app:proxy:csca}

Table~\ref{tab:incremental-csca} reports the CS/CA-stratified incremental proxy validity results referenced in \S\ref{sec:proxy:incremental}.

\begin{table}[htbp]
  \centering
  \setlength{\tabcolsep}{0.25em}
  \renewcommand{\arraystretch}{1.05}
  \begin{tabular}{@{}llcccc@{}}
    \toprule
    & & \multicolumn{2}{c}{\textbf{CA items}} & \multicolumn{2}{c}{\textbf{CS items}} \\
    \cmidrule(lr){3-4} \cmidrule(lr){5-6}
    \textbf{Native} & \textbf{L} & $\boldsymbol{\Delta R^2}$ & $\boldsymbol{p}$ & $\boldsymbol{\Delta R^2}$ & $\boldsymbol{p}$ \\
    \midrule
    Skolprov    & SV & .140 & $<$10\textsuperscript{-13} & .060 & $<$10\textsuperscript{-3} \\
    INCLUDE     & LT & .095 & $<$10\textsuperscript{-5}  & .071 & $<$10\textsuperscript{-7} \\
    INCLUDE     & ET & .059 & .003             & .046 & .010 \\
    NRK-Quiz    & NB & .048 & $<$10\textsuperscript{-4}  & .044 & $<$10\textsuperscript{-7} \\
    Exam-ET     & ET & .031 & $<$10\textsuperscript{-3}  & .015 & $<$10\textsuperscript{-4} \\
    DanishCT    & DA & .005 & .333             & .014 & .005 \\
    SwedishFacts& SV & .007 & .573             & .046 & .180 \\
    \bottomrule
  \end{tabular}
  \caption{CS/CA-stratified incremental proxy validity. $\Delta R^2$ and $p$ from the nested $F$-test, computed separately on CA and CS items of each native benchmark. Proxy scores use the corresponding CA/CS split of G-MMLUX.}
  \label{tab:incremental-csca}
\end{table}

\subsection{Incremental Proxy Validity (Details)}
\label{sec:app:proxy:incremental}

The main analysis fits nested OLS models on the 29 evaluated models.
For each pairing, the baseline model uses $\log(\text{total parameters})$ and same-language Belebele.
The extended model adds the translated MMLU proxy score.
We quantify uncertainty by resampling the 29 models with replacement 5{,}000 times and refitting both models in every resample.
We report 95\% percentile bootstrap CIs for $R^2$, $\Delta R^2$, and coefficients.
The final reported outputs are reproduced with bootstrap seed~42; this seed is distinct from the Native-500 IAA bootstrap seed reported in Appendix~\ref{sec:app:annotation:iaa}.

A pairing is classified as \emph{robust} if (i)~the full-sample nested $F$-test is significant, (ii)~the 95\% model-bootstrap CI for $\beta_\text{proxy}$ excludes zero, and (iii)~$\beta_\text{proxy}$ is positive with nominal $p<.05$ in all 17 LOFO refits.
A \emph{marginal} pairing is full-sample significant but fails at least one robustness criterion.
A \emph{null} pairing has a non-significant full-sample $F$-test and a bootstrap $\beta_\text{proxy}$ CI containing zero.
The labels describe evidence strength, not benchmark quality. Because in-sample $\Delta R^2$ is non-negative by construction, its percentile-bootstrap CI cannot serve as a null test; near-zero lower bounds describe uncertainty in effect magnitude but do not independently determine the class.

The CS/CA-stratified regressions in Table~\ref{tab:incremental-csca} are exploratory secondary analyses and do not determine the robust/marginal/null class.
For Exam-ET, the small $\Delta R^2=.024$ occurs against a high baseline ($R^2=.954$), so the translated proxy adds a modest but robust increment after controls.
For SwedishFacts, the low baseline ($R^2=.353$) indicates that model size and Belebele explain substantially less variance than for the exam-style pairings; the data do not by themselves identify this residual variance as a uniquely local construct.

\paragraph{OLS assumptions and collinearity.}
With $n=29$ and three predictors in the extended main model, the regression has limited degrees of freedom, motivating the model-level bootstrap and LOFO analyses.
Variance Inflation Factors indicate substantial overlap among model strength, language proficiency, and proxy performance; this inflates coefficient uncertainty and reinforces interpreting $\beta_\text{proxy}$ as approximate within the evaluated model set.
The nested $F$-test and $\Delta R^2$ quantify whether the added proxy improves the fitted model, while the bootstrap and LOFO analyses assess how stable that conclusion is to model sampling. In particular, because in-sample $\Delta R^2\geq 0$ by construction, a percentile-bootstrap interval whose lower endpoint rounds to .000 (or is slightly above zero) should not be interpreted as a separate rejection of the null.

\subsection{Robustness (Details)}
\label{sec:app:proxy:robustness}

\paragraph{Partial correlations.}
Table~\ref{tab:app:partial-corr} reports the partial correlation between native and proxy performance after residualizing both on the main controls ($\log(\text{total parameters})+$ Belebele).
The partial-correlation pattern mirrors the main incremental results: 
robust pairings retain strong residual native--proxy associations, whereas SwedishFacts remains weak and non-significant after the same controls.

\begin{table}[htbp]
\centering
\setlength{\tabcolsep}{0.3em}
\begin{tabular}{@{}llccr@{}}
\toprule
\textbf{Native} & \textbf{L} & \textbf{Sub.} & $\boldsymbol{r_\text{partial}}$ & $\boldsymbol{p}$ \\
\midrule
Skolprov    & SV & ALL & .943 & $<10^{-14}$ \\
            &    & CA  & .945 & $<10^{-14}$ \\
            &    & CS  & .670 & $<10^{-4}$ \\
INCLUDE     & LT & ALL & .828 & $<10^{-7}$ \\
            &    & CA  & .747 & $<10^{-5}$ \\
            &    & CS  & .825 & $<10^{-7}$ \\
NRK-Quiz    & NB & ALL & .785 & $<10^{-6}$ \\
            &    & CA  & .716 & $<10^{-4}$ \\
            &    & CS  & .823 & $<10^{-7}$ \\
Exam-ET     & ET & ALL & .730 & $<10^{-5}$ \\
            &    & CA  & .661 & $<10^{-4}$ \\
            &    & CS  & .718 & $<10^{-5}$ \\
INCLUDE     & ET & ALL & .570 & .001 \\
            &    & CA  & .545 & .002 \\
            &    & CS  & .485 & .008 \\
DanishCT    & DA & ALL & .426 & .021 \\
            &    & CA  & $-$.194 & .314 \\
            &    & CS  & .520 & .004 \\
\midrule
SwedishFacts& SV & ALL & .233 & .223 \\
            &    & CA  & .113 & .558 \\
            &    & CS  & .266 & .163 \\
\bottomrule
\end{tabular}
\caption{
Partial correlations between native and proxy scores after residualizing on the main controls. $n=29$ for all entries.
}
\label{tab:app:partial-corr}
\end{table}

\begin{table*}[t]
\centering
\setlength{\tabcolsep}{0.95em}
\begin{tabular}{@{}llrrrrc@{}}
\toprule
\textbf{Native} & \textbf{L} & $\boldsymbol{R^2_\text{base}}$ & $\boldsymbol{R^2_\text{ext}}$ & $\boldsymbol{\Delta R^2}$ \textbf{[95\% CI]} & $\boldsymbol{\beta_\text{prx}}$ \textbf{[95\% CI]} & $\boldsymbol{p_F}$ \\
\midrule
Skolprov       & SV & .882 & .987 & .105 [.018,.224] & 1.166 [.883,1.285] & $<10^{-12}$ \\
INCLUDE        & LT & .890 & .977 & .087 [.024,.160] & 1.082 [.843,1.298] & $<10^{-9}$ \\
INCLUDE        & ET & .846 & .934 & .087 [.006,.233] & 1.057 [.365,1.418] & $<10^{-5}$ \\
NRK-Quiz       & NB & .940 & .977 & .037 [.006,.080] & 0.537 [.342,.679] & $<10^{-6}$ \\
Exam-ET        & ET & .956 & .986 & .030 [.013,.051] & 0.587 [.393,.947] & $<10^{-7}$ \\
DanishCT       & DA & .952 & .971 & .020 [.003,.045] & 0.572 [.340,.916] & $<10^{-3}$ \\
SwedishFacts   & SV & .356 & .430 & .073 [.004,.187] & 0.247 [.059,.476] & .092 \\
\bottomrule
\end{tabular}
\caption{
Sensitivity analysis with English G-MMLUX as an additional control. 
Brackets show 95\% percentile model-bootstrap CIs. 
This specification is a robustness check against general MMLU-content familiarity and is not used to define the main robust/marginal/null evidence classes.
}
\label{tab:app:sensitivity-full}
\end{table*}

\paragraph{Leave-one-family-out sensitivity.}
LOFO removes every model family in turn, including singleton families, and refits the ALL-items main model.
There are 17 excluded families and 119 total refits (7 pairings $\times$ 17 families); each refit retains 25--28 of the 29 models.
Table~\ref{tab:app:lofo-summary} summarizes all 17 exclusions per pairing.

\begin{table}[htbp]
\centering
\setlength{\tabcolsep}{0.25em}
\renewcommand{\arraystretch}{1.08}
\begin{tabular}{@{}lcll@{}}
\toprule
\textbf{Pairing} & \textbf{Sig.} & \textbf{Drops} & \textbf{Class} \\
\midrule
Skolprov     & 17/17 & none & robust \\
INCLUDE-LT   & 17/17 & none & robust \\
NRK-Quiz-QA  & 17/17 & none & robust \\
Exam-ET      & 17/17 & none & robust \\
INCLUDE-ET   & 16/17 & Qwen (.102) & marginal \\
DanishCT     & 14/17 & \makecell[l]{Cohere (.110);\\ Llama (.072);\\ Qwen (.096)} & marginal \\
SwedishFacts & 0/17  & all 17 & null \\
\bottomrule
\end{tabular}
\caption{
LOFO stability across all 17 model-family exclusions.
\textbf{Sig.} reports how many refits retain a positive $\beta_\text{proxy}$ with nominal $p<.05$.
\textbf{Drops} lists the family exclusions for which this criterion is not met.
These nominal tests are not multiplicity-corrected and are used only as a conservative stability requirement for the \emph{robust} label.
}
\label{tab:app:lofo-summary}
\end{table}

The four robust pairings preserve their direction and nominal significance under every family exclusion.
DanishCT and INCLUDE-ET remain positive in the full sample but are sensitive to changes in the evaluated model set.
SwedishFacts remains non-significant throughout, supporting the stability of the null classification within this model sample.

\subsection{Sensitivity: English MMLU Control (Details)}
\label{sec:app:proxy:sensitivity}

The sensitivity analysis adds English G-MMLUX performance as a third baseline control.
This checks whether the translated-proxy association mainly reflects general familiarity with MMLU content.
Let $C_L$ denote the three controls $\log(\text{total parameters})$, $\text{Belebele}_L$, and GMMLUX-EN.
We compare $\text{native}_L\sim C_L$ with $\text{native}_L\sim C_L+\text{proxy}_L$.
As in the main analysis, confidence intervals use 5{,}000 model-level bootstrap resamples with seed~42.

Six pairings remain significant by the nested $F$-test.
SwedishFacts remains non-significant by the $F$-test ($p=.092$), but the bootstrap CI for $\beta_\text{prx}$ is positive ([.059,.476]).
We therefore treat SwedishFacts as weak mixed positive evidence in this alternative specification, not as a clean null.
This does not change its \emph{null} classification under the fixed main-specification evidence definition.

The sensitivity analysis preserves the broad ordering after controlling for English MMLU performance.
DanishCT shows a stronger signal in this specification, but its main evidence class remains \emph{marginal} because the primary classification is based on the main specification and LOFO/bootstrap criteria.
The sensitivity should not be read as identifying a purely language-specific causal effect, since English MMLU, Belebele, and the translated proxy are correlated measures of overlapping model abilities.

\end{document}